\documentclass{article}
\usepackage[preprint]{neurips_2026}
\usepackage[utf8]{inputenc}
\usepackage[T1]{fontenc}
\usepackage{url}
\usepackage{booktabs}
\usepackage{array}
\usepackage{amsmath}
\usepackage{amsfonts}
\usepackage{amssymb}
\usepackage{nicefrac}
\usepackage{microtype}
\usepackage{xcolor}
\usepackage{graphicx}
\usepackage{float}
\usepackage{placeins}
\usepackage[hidelinks]{hyperref}

\graphicspath{{figures/}}

\title{How Alignment Routes: Localizing, Scaling, and\\Controlling Policy Circuits in Language Models}

\author{
  Greg Frank \\
  Chief Scientist, MACE AI \\
  \texttt{greg@mace.ai}
}

\makeatletter
\renewcommand{\@noticestring}{Accepted at the Mechanistic Interpretability Workshop at the 43rd International Conference on Machine Learning (ICML), 2026.}
\renewcommand{\@notice}{%
  \enlargethispage{2\baselineskip}%
  \@float{noticebox}[b]%
    \vspace*{14pt}%
    \fontsize{7.5pt}{9pt}\selectfont\noindent\mbox{\@noticestring}%
  \end@float%
}
\makeatother

\begin{document}
\maketitle

\begin{abstract}
Models can represent policy-relevant content yet respond differently.
We study the policy-routing circuit that turns detection into refusal.
Using political censorship and harmful-request refusal as test cases, four-method circuit analyses in Qwen3-8B, Phi-4-mini, and Gemma-2-2B reveal sparse gate--amplifier circuits at architecture-specific locations.
An intermediate-layer gate initiates routing, while later, distributed amplifier heads strengthen a refusal-directed signal.
The gate is hard to find with direct logit attribution (DLA): in Qwen3-8B, it contributes less than 1\% of output DLA.
We therefore combine DLA screening, directional intervention, activation interchange, and gate-to-amplifier knockout.
Activation interchange identifies upstream trigger candidates, and knockout suppresses a majority of tested later-head contributions in all three models.
The gate is an upstream trigger, not a self-contained refusal mechanism; later amplifiers strengthen the signal toward the output.
Candidate-head screens in nine additional models from six labs, spanning 2B--72B parameters, nominate similarly localized routing candidates without establishing head-level homology.
Policy routing is a sparse, causally structured interface between content detection and aligned behavior that output attribution alone can miss.
\end{abstract}

\noindent\textbf{Code and data:} \url{https://github.com/gregfrank/how-alignment-routes}

\section{Introduction}
\label{sec:intro}

Models can strongly represent policy-relevant content without adopting the same response policy.
In a companion study, political sensitivity was highly linearly decodable at mid-depth in four models whose Chinese-language outputs differed markedly among propaganda, evasion, accurate answers, and refusal \citep{frank2026detection}.
Detection alone therefore does not explain which response a model selects.

We ask a causal question: given that policy-relevant content is represented, which components help turn that representation into refusal?
We call this computation \emph{policy routing}.
The distinction between harmfulness and refusal representations established by \citet{zhao2025harmfulness} motivates rather than conflicts with this question: their result separates the representations; ours asks which components map one into the other.
Related safety-head studies localize sparse components \citep{zhou2025safetyheads,chu2025keyheads}; we test their ordering and dependence within a circuit.

Our answer is a sparse attention interface embedded in broader computation.
An intermediate-layer \emph{gate} initiates routing, and later \emph{amplifier} heads strengthen a refusal-directed signal.
The gate is not a self-contained policy switch: it contributes little to the final output attribution and cannot produce refusal without downstream computation.
In Qwen3-8B, the strongest tested entry-trigger signature concentrates at L17.H17, while later heads form a distributed cascade.
Phi-4-mini and Gemma-2-2B show the same functional ordering at architecture-specific locations.

\begin{figure}[H]
\centering
\includegraphics[width=\textwidth]{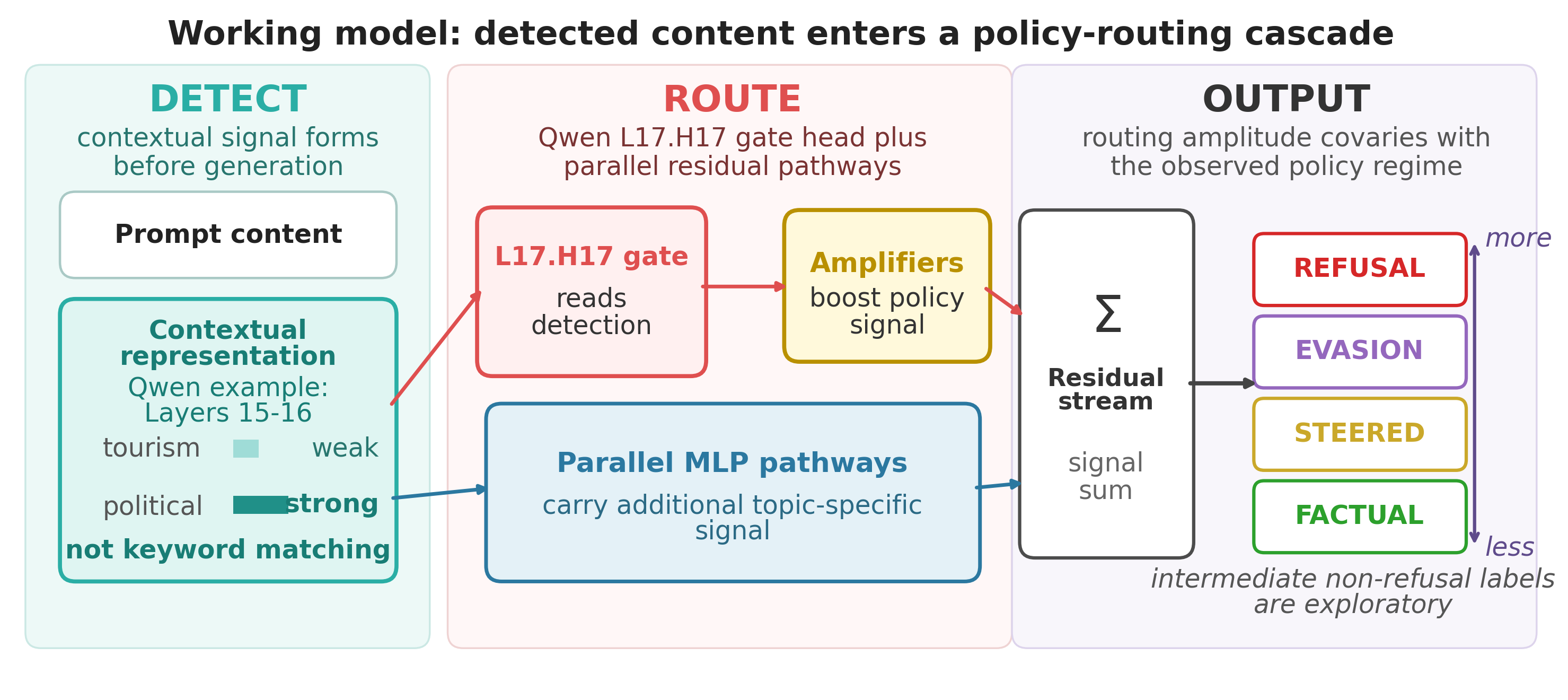}
\caption{%
\textbf{Qwen3-8B routing-mechanism schematic.}
The measured detection signal forms at layers~15--16.
A gate head writes a routing vector; amplifier heads boost it toward refusal.
MLP pathways carry topic-specific signal in parallel.
Detection-layer modulation changes binary refusal; the finer non-refusal categories shown here remain exploratory.%
}
\label{fig:overview}
\end{figure}

We identify this organization by combining output attribution, directional intervention, activation interchange, and gate-to-amplifier knockout.
The four-method circuit analysis is performed in Qwen, Phi, and Gemma; nine additional models receive candidate-head screens without knockout confirmation.
We then test generated behavior separately, including a human-scored Phi intervention, and use a substitution cipher as a bounded stress test of the routing interface.
The experiments cover Chinese-political censorship and harmful-request refusal; they do not establish the same organization for every alignment behavior.

\section{Problem setup and evidence scope}
\label{sec:detection}

\subsection{Detection does not determine policy}
\label{sec:puzzle}

The companion study examined 128 human-reviewed Chinese-language outputs.
Qwen2.5-7B produced propaganda in 94\% of cases and accurate answers in 3\%; DeepSeek-R1 produced propaganda in 78\% and evasion in 41\%; GLM-4-9B produced accurate answers in 88\% and refusals in 12\%; and Phi-4-mini produced evasion in 59\% and accurate answers in 34\% \citep{frank2026detection}.
Labels were nonexclusive, and no model was refusal-dominant.
Despite these different output distributions, best-layer held-one-category-out linear-probe accuracy was 93.8--99.0\% in all four models.
Across all four models, political sensitivity is strongly linearly decodable at mid-depth.
The shared property is linear decodability, not a common direction, geometry, or code.

Probe accuracy requires controls.
Political probes can reach 100\% accuracy, but so can null probes trained on arbitrary label-shuffled splits \citep{hewitt2019designing}.
Leave-one-category-out cross-validation, in which the probe trains on all but one political category and tests on the held-out category, separates the two: political probes retain 91--100\% accuracy while null probes fall to chance.

In Qwen3-8B, the per-layer logit-DLA profiles at the last prompt token and first generated token overlap closely (Appendix Figure~\ref{fig:contextual}, panel A).
The close match shows that the measured refusal-directed profile is already present during prompt processing; by itself, it does not identify the component that controls the response.
Even GLM-4-9B, which produced refusals on only 12\% of the companion behavioral sample, shows a 2.8-nat KL peak between matched sensitive and control prompts in the contextual-detection analysis~\citep{frank2026detection}.

Detection is compositional: the same keyword produces different layer-16 scores depending on framing, and routing depends on more than a scalar threshold (Appendix Figure~\ref{fig:contextual}, panel B).

Together, these results show that the measured signal is present before generation and varies with context.
They do not identify the components that route it into a response.

\subsection{What each measurement establishes}
\label{sec:endpoints}

Unless stated otherwise, $n$ counts matched sensitive/control prompt pairs; generated-output counts are labeled separately.
A \emph{routing signal} is a scalar readout along a named direction, not a direction-independent hidden variable.
Output \emph{logit direct attribution} (logit-DLA) projects a component's residual contribution onto a refusal-minus-matched-answer unembedding direction and reports the sensitive-minus-control difference.
\emph{Probe-direction attribution} instead projects a head or MLP output onto a frozen sensitive-minus-control residual-space direction.
The \emph{MLP pathway} refers to the feed-forward blocks' direct residual contributions under the stated projection.
The intervention endpoints below use teacher-forced matched-continuation negative log-likelihood (NLL).

A \emph{gate} is an upstream component or sparse locus whose state influences whether a downstream cascade begins; it need not contain the policy or determine the response by itself.
\emph{Amplifiers} are later heads that strengthen the refusal-directed signal.
We use \emph{four-method circuit analysis} for the prespecified combination of logit-DLA, directional intervention, activation interchange, and knockout cascade.
It is not an exhaustive decomposition of all model computation.

The methods answer different questions.
Logit-DLA screens for output carriers; directional intervention and activation interchange test causal influence under named endpoints.
Knockout tests directed dependence on later heads, and generation with human labels measures final-output behavior.
Appendix Table~\ref{tab:evidence} maps each measurement to the claim it supports.

\subsection{Models, domains, and evidence tiers}
\label{sec:scope}

Qwen3-8B, Phi-4-mini, and Gemma-2-2B receive the four-method analysis, including a gate-to-amplifier knockout test.
The remaining nine models receive logit-DLA, intervention, and interchange screens; these screens nominate candidate loci but do not establish a complete shared mechanism or head-level homology.
Generated behavior is a separate evidence tier: we test head-group interventions in Phi and Gemma and detection-layer modulation in Qwen and Phi.
The cipher study uses a second, explicitly distinguished interchange endpoint.
Appendix~\ref{app:evidence} gives the complete claim-to-evidence map.

\FloatBarrier

\section{Qwen3-8B: A localized trigger in a distributed cascade}
\label{sec:circuit}

\subsection{Output attribution finds carriers but misses the gate}
\label{sec:pipeline}

We first decompose the paired refusal-minus-answer attribution across all 1,152 Qwen attention heads.
Deep output carriers dominate this screen: on the 24-pair discovery corpus, L35.H25 is the top head and L17.H17 ranks 149th by absolute output logit-DLA; in the $n{=}120$ confirmation, L17.H17 ranks 292nd.
The DLA top-10 Jaccard under 2,000 bootstrap resamples is 0.66, showing that the extreme attribution ranking is sampling-sensitive on this corpus.
Logit-DLA supplies candidates; it does not establish causal control.

\subsection{Directional intervention and activation interchange}
\label{sec:qwen_interchange}

No single screen identifies the gate.
We therefore narrow candidates by directional intervention and then test their prompt-conditioned activations by interchange.

\paragraph{Directional head intervention.}
For each candidate, we attenuate a frozen residual-space detection direction in that head's activation and measure the resulting change in the teacher-forced NLL of the matched-control continuation.
Layers~22--23 dominate this ranking on the discovery corpus.
L22.H7 is the strongest individual head, reducing the baseline gap by 8.8\%, while L17.H17 ranks sixth at 1.8\%.
The intervention top-10 bootstrap Jaccard is 0.92, with a fifth percentile of 0.82.

\paragraph{Activation interchange.}
A directional intervention tests whether attenuating the specified head-space direction changes matched-continuation NLL; interchange tests whether swapping the prompt-conditioned activation changes that same endpoint.
For each head, we swap its activation between a sensitive and a matched control prompt at all overlapping prompt positions (Appendix~\ref{app:methods}).
The \emph{necessity} test runs on a sensitive prompt but replaces one head's activation with what it produces on a matched control; if this shrinks the matched-continuation NLL gap, the head carried prompt-specific information relevant to that endpoint.
The \emph{sufficiency} test runs on a control prompt but injects one head's activation from the sensitive prompt; a positive effect means that this activation shifts the matched-continuation NLL endpoint in the routing-consistent direction.
We call joint positive necessity and sufficiency an operational \emph{trigger signature}; it does not show that the activation alone determines generated output.
Heads with necessity but little sufficiency are operational amplifier candidates: they strengthen the tested endpoint but do not independently initiate it under interchange.

\begin{figure}[tbp]
\centering
\includegraphics[width=\textwidth]{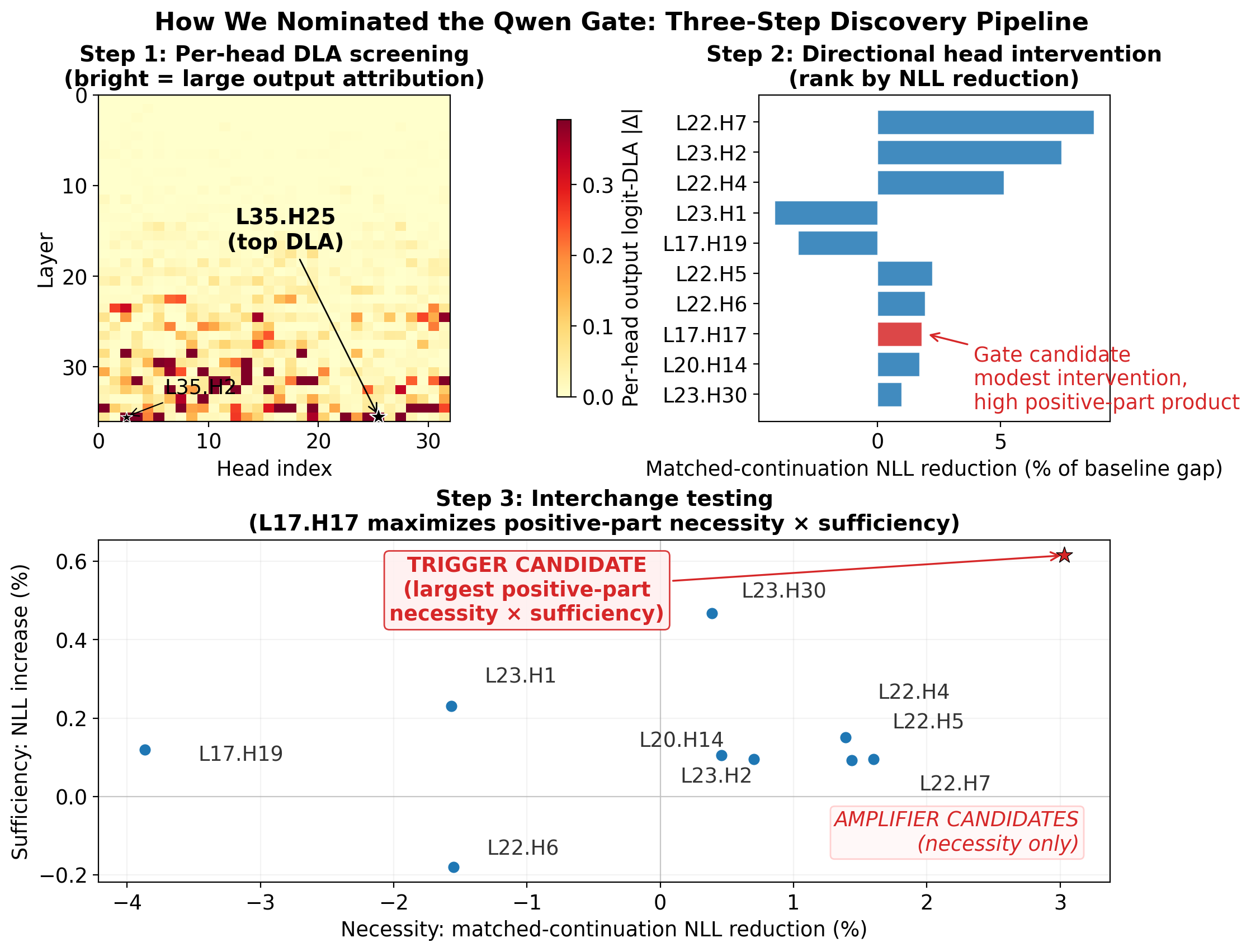}
\caption{%
\textbf{Three-step Qwen3-8B discovery pipeline.}
The plotted discovery results use 24 matched pairs; the null-calibrated interchange confirmation reported in the text uses 120 pairs.
\textbf{Step 1:} Per-head output logit-DLA heatmap; deep layers dominate.
\textbf{Step 2:} Directional head intervention scored by matched-control-continuation NLL; layers~22--23 dominate, L22.H7 leads, and L17.H17 is sixth.
\textbf{Step 3:} NLL-scored necessity $\times$ sufficiency; L17.H17 has the strongest joint signature among the ten candidates tested.%
}
\label{fig:discovery}
\end{figure}

The interchange set comprised the ten highest-ranked directional-intervention candidates on the discovery corpus; the directional stage followed the all-head output-DLA screen, and interchange was not run over all 1,152 heads.
Among the ten Qwen candidates tested, L17.H17 has the strongest joint interchange signature: 1.127\% necessity and 0.336\% sufficiency at $n{=}120$.
For the joint statistic $T_h=[\overline{N}_h]_+[\overline{S}_h]_+$, its raw value is $9.18\times10^{-4}$, 69.6 times the paired sign-flip null mean.
No one of 10,000 null draws matched or exceeded this value either for L17.H17 or for the maximum across the ten tested heads (add-one Monte Carlo $p\approx10^{-4}$ for both comparisons).
The absolute change is modest, but its joint necessity--sufficiency evidence marks L17.H17 as the strongest upstream entry-trigger candidate among those ten (Figure~\ref{fig:discovery}).
DLA screening, directional intervention, and NLL-scored interchange produce different rankings; their convergence nominates an upstream candidate distinct from the downstream carriers.

The core amplifier heads (L22.H7, L23.H2, L22.H4) remain the top three directional-intervention heads on broader corpora of 32 and 120~pairs.
Approximately half of heads at directional-intervention ranks~7--20 vary with corpus composition.

\subsection{Functional localization and downstream dependence}
\label{sec:roles}

The gate head's (L17.H17) attention pattern is consistent with sensitivity to the relevant content.
On politically sensitive prompts, its attention concentrates on the relevant token; on matched controls with identical syntax, it attends to generic punctuation.
The gate sits at layer~17, after the detection signal has formed at layers~15--16.

The tested amplifier heads' measured attention is dominated by formatting and position tokens rather than the content tokens highlighted by the gate.
They boost the routing signal written upstream.

At $n{=}120$, knocking out L17.H17 reduces the ordinary output logit-DLA of five of six tested later amplifier heads, with pointwise 95\% paired-bootstrap intervals below zero.
For example, Qwen L22.H7 changes by $-0.0280$ with interval $[-0.0365,-0.0199]$.
This partial suppression is evidence of distributed downstream dependence, not of a gate that independently determines the generated response.
Section~\ref{sec:replication} compares the corresponding tests in Phi and Gemma, and Appendix~\ref{app:cascade_intervals} reports every interval.

\subsection{The gate is a trigger, not a carrier}
\label{sec:mlp}

Output logit-DLA at $n{=}120$ reveals a seeming paradox: L17.H17 and the tested amplifier heads each contribute less than 1\% directly, yet NLL-scored interchange separates L17.H17 from its null and gate knockout suppresses later-head logit-DLA contributions.
In the prefix-restricted probe-direction diagnostic, L17.H17 has the second-largest absolute mean contribution among head outputs written through L17.
The rank is not a temporal trace of a fixed activation; it locates the head's write relative to the heads available at that cutoff.
At the output, other distributed carriers dominate ordinary logit-DLA.
The three methods resolve this mismatch.
Output attribution finds carriers but misses the gate.
Interchange identifies L17.H17 as the strongest tested entry trigger, and knockout shows that later contributions depend on it.
The gate initiates the measured cascade rather than carrying most of its final output signal.

Attention is not the whole computation.
MLP blocks contribute about 23\% of the direct attribution on the diverse $n{=}120$ corpus and more on concentrated single-topic prompts.
Separate probe-direction interchange and knockout diagnostics find large MLP effects while retaining a measurable L17.H17 effect, but do not yet decompose the MLP pathway at the feature level (Appendix~\ref{app:methods}).

\section{Architecture-specific replication and scaling}
\label{sec:crossarch}

\subsection{Four-method analyses in Phi and Gemma}
\label{sec:replication}

Phi-4-mini and Gemma-2-2B provide two additional four-method analyses.
In Phi-4-mini, L13.H7 is both the strongest directional-intervention head (1.422 nats/token) and the strongest joint interchange candidate among the tested set, with necessity and sufficiency equal to 3.37\% and 3.31\% of the baseline NLL gap.
In Gemma-2-2B, L13.H2 likewise leads both tests: its directional-intervention effect is 1.015 nats/token, with 8.45\% necessity and 1.34\% sufficiency.
The coincidence between the directional and interchange candidates in these models differs from Qwen, where L22.H7 leads the directional test while L17.H17 has the strongest joint trigger signature.

\begin{table}[H]
\caption{\textbf{Four-method comparison in the three deeply analyzed models ($n{=}120$).} Necessity and sufficiency are percentages of the baseline matched-continuation NLL gap. The directional column gives the top head and raw NLL reduction in nats/token. The final column counts tested heads later than the gate whose output logit-DLA is suppressed after knockout.}
\label{tab:deep_models}
\centering
\small
\begin{tabular}{@{}llllc@{}}
\toprule
Model & Gate candidate & Nec. / suff. & Top directional (NLL) & Later-head suppression \\
\midrule
Qwen3-8B & L17.H17 & 1.13\% / 0.34\% & L22.H7 (0.137) & 5/6 \\
Phi-4-mini & L13.H7 & 3.37\% / 3.31\% & L13.H7 (1.422) & 3/5 \\
Gemma-2-2B & L13.H2 & 8.45\% / 1.34\% & L13.H2 (1.015) & 2/3 \\
\bottomrule
\end{tabular}
\end{table}

\begin{figure}[tbp]
\centering
\includegraphics[width=\textwidth]{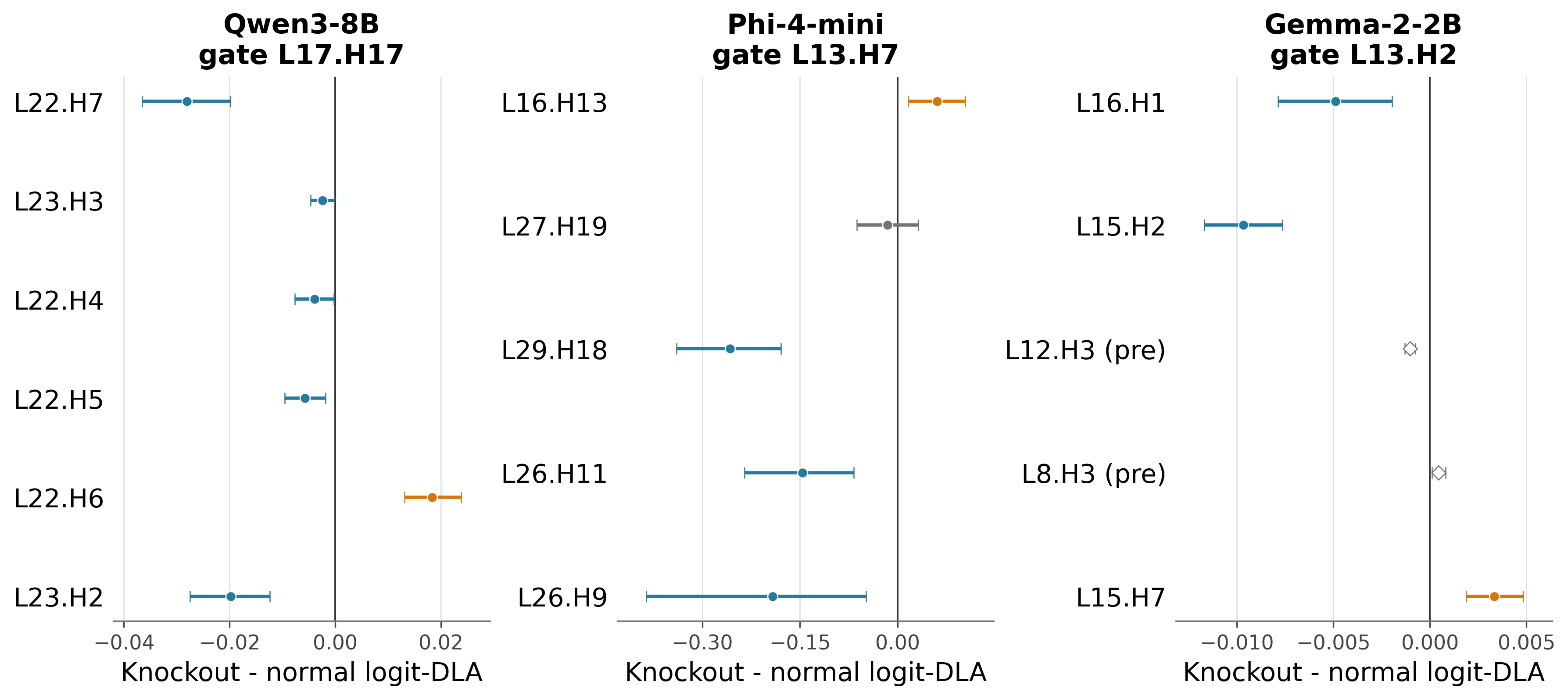}
\caption{%
\textbf{Gate-knockout changes in selected-head output logit-DLA ($n{=}120$).}
Points show the raw change in each selected head's ordinary output logit-DLA after gate knockout; bars are pointwise 95\% paired-bootstrap intervals over matched prompt pairs.
Negative values indicate suppression.
Intervals fall below zero for 5/6 later Qwen heads, 3/5 later Phi heads, and 2/3 later Gemma heads.
Two additional pre-gate Gemma comparison heads are displayed but do not test downstream dependence.%
}
\label{fig:knockout}
\end{figure}

Gate knockout then tests whether these candidates affect later components.
The pointwise 95\% intervals fall below zero for five of six tested later heads in Qwen, three of five in Phi, and two of three in Gemma (Figure~\ref{fig:knockout}).
Representative raw knockout-minus-normal intervals are $[-0.0365,-0.0199]$ for Qwen L22.H7, $[-0.339,-0.180]$ for Phi L29.H18, and $[-0.0117,-0.0076]$ for Gemma L15.H2; Phi L27.H19 is inconclusive.
The prespecified Gemma comparison set also included L12.H3 and L8.H3, which precede the L13 gate.
Their output-normalized DLA changes are shown for completeness but are not evidence of downstream dependence.
Across the genuinely later heads, partial rather than uniform suppression points to architecture-specific, distributed cascades.

\FloatBarrier

\subsection{Internal replication does not imply uniform behavioral effects}
\label{sec:behavior_bridge}

Functional ordering at an internal endpoint does not guarantee that every local ablation will change a sampled response.
In an initial automated Phi generation screen ($n{=}30$ per condition), refusal was 30/30 at baseline and after L13.H7 ablation, 28/30 after the L13 band, and 18/30 after a 26-head cross-layer intervention.
The corresponding Gemma all-band intervention changed refusal only from 30/30 to 29/30, a negative endpoint despite the internal dependence above.
Section~\ref{sec:control} tests final-output behavior directly with human labels and preserves both the positive cross-layer result and the inconclusive local effects.

\subsection{Candidate-head screens in nine additional models}
\label{sec:panel}

For the cross-model panel, the interchange screen ranks a fixed candidate set by signed mean necessity plus signed mean sufficiency; unlike the Qwen positive-part product, this selection score does not require both terms to be positive.
Applied to nine additional models, these screens identify localized, necessity-dominant candidate loci.
They neither establish Qwen-like trigger signatures nor imply head-level homology or a shared complete mechanism (full table in Appendix~\ref{app:scaling}).
Selected-head necessity ranges from 1.0\% (Mistral-7B) to 4.7\% (GLM-Z1-9B) across these nine models; both 70B+ models retain positive necessity at the largest scales tested.

\subsection{Pipeline-scoped scaling}
\label{sec:scaling}

Four same-generation pairs provide a pipeline-matched scaling comparison under a fixed candidate-selection and intervention pipeline (Appendix Figure~\ref{fig:scaling}; per-model details in Appendix~\ref{app:scaling}).
The top per-head directional-intervention effect is smaller at the larger checkpoint in every pair: about 8$\times$ in Gemma-2, 1.3$\times$ in Qwen3, 17$\times$ in Phi-4, and 58.1$\times$ in Qwen2.5.
At Qwen2.5-72B, the top NLL reduction is 0.01558 nats/token.
Panel-selected interchange necessity remains above 1\% in all cases, including the largest model tested (72B).
Candidate locations and cross-layer dispersion vary by architecture.
Among the tested methods, interchange therefore remains informative as single-head directional-intervention effects weaken.
Candidate selection and the intervened residual-space direction remain pipeline choices, so this panel does not establish invariance to alternative target definitions.

\section{From internal circuits to generated behavior}
\label{sec:control}

\subsection{Behavioral endpoints}
\label{sec:behavior_endpoints}

Internal attribution, NLL, and knockout effects establish causal control of specified internal endpoints.
Whether an intervention changes a generated response is a separate question.
We evaluate binary refusal directly and use human labels where the claims depend on output behavior.

\subsection{Human-scored Phi head-group interventions}
\label{sec:band_behavior}

For the Phi test, we generated 720 outputs from 120 safety prompts under six prespecified intervention conditions and obtained human binary refusal labels for a balanced 198-response subset (33 per condition).
Two condition-blinded annotators agreed on 194/198 responses (Cohen's $\kappa=0.948$); a third blinded annotator adjudicated four disagreements.

\begin{table}[H]
\caption{%
\textbf{Human-scored Phi-4-mini head-group interventions.}
Differences are relative to the 33/33 baseline, with unadjusted Newcombe hybrid-score 95\% intervals; Fisher exact $p$-values are Holm-adjusted across the five contrasts.%
}
\label{tab:phi_band_behavior}
\centering
\small
\begin{tabular}{@{}lrrr@{}}
\toprule
Condition & Refusal & Difference (pp) & Holm $p$ \\
\midrule
Baseline & 33/33 & --- & --- \\
Single L13.H7 & 32/33 & $-3.0$ [$-15.3$, $+7.7$] & 1.0 \\
L13 band (5 heads) & 30/33 & $-9.1$ [$-23.6$, $+2.9$] & 0.72 \\
L16 band (7 heads) & 30/33 & $-9.1$ [$-23.6$, $+2.9$] & 0.72 \\
Cross-layer top-10 & 8/33 & $-75.8$ [$-87.2$, $-56.0$] & $1.32\!\times\!10^{-10}$ \\
Legacy mixed 26-head group & 14/33 & $-57.6$ [$-72.8$, $-37.8$] & $3.79\!\times\!10^{-7}$ \\
\bottomrule
\end{tabular}
\end{table}

Table~\ref{tab:phi_band_behavior} shows that the two cross-layer interventions produce large reductions in refusal, whereas the single-head and individual within-layer estimates remain inconclusive at this sample size.
An exact sensitivity analysis on prompts shared with baseline yields the same pattern: cross-layer top-10 changes 11/15 responses from refusal to non-refusal with no reversals (Holm-adjusted McNemar $p=0.00391$), and the legacy group changes 15/25 with no reversals ($p=0.000305$).
This supplies direct final-output evidence for distributed cross-layer control in Phi.
It does not retest the separate Qwen L17.H17 localization.

We selected the scored subset post hoc without consulting label values, response text, automated scores, or routing measurements.
It is a value-blind validation subset rather than a preregistered endpoint.
Appendix~\ref{app:behavioral_validation} gives the selection procedure and a separate association analysis; the latter does not validate the residual projection as a prompt-level behavioral surrogate.

\subsection{Detection-layer dose response}
\label{sec:dose}

In the companion study, residual-stream ablation of the political-sensitivity direction changed output routing in three of four models, while cross-model direction transfer failed \citep{frank2026detection}.
Here we test the same intervention family over a range of magnitudes and evaluate generated behavior.
By adding or subtracting a scaled steering direction ($\alpha \cdot d$, where $d$ is the mean activation difference between sensitive and control prompts) at the detection layer, we vary refusal rates across intervention doses.
Three model judges assign six-way descriptive labels to 2,400 outputs per model at $n{=}120$, but our calibrated quantitative endpoint is binary refusal versus non-refusal.

On Tiananmen prompts (the only category with 100\% baseline refusal), the automated binary-refusal curve falls monotonically from 100\% to 0\% by $\alpha{=}35$ (Appendix Figure~\ref{fig:dose_response}, panel A).
The broader $n{=}120$ corpus reveals that refusal is topic-specific.
Across 15 political categories at $\alpha{=}0$, only Tiananmen triggers consistent hard refusal (8/8); the aggregate refusal rate is 9/120 (7.5\%), masking the topic-specific structure (Appendix~\ref{app:percategory}).
The automated binary endpoint reveals different refusal thresholds across categories under amplification (Appendix Figure~\ref{fig:dose_response}, panel B).

In a category-stratified audit of 120 outputs, two condition-blinded annotators agreed on 107/120 labels (Cohen's $\kappa=0.779$), and a third annotator resolved 13 disagreements.
Among covered machine-majority strata, post-stratified binary accuracy was 96.8\% for Qwen (2,331/2,400 outputs covered) and 88.3\% for Phi (2,349/2,400 covered).
The automated mapping undercounted aggregate refusal, especially in Phi, and the audit does not validate individual dose contrasts or the six-way taxonomy (Appendix~\ref{app:behavioral_validation}).

\subsection{Exploratory output composition}
\label{sec:composition}

The machine-assigned categories suggest that attenuation can replace refusal with evasive or factual output on the political corpus and with non-refusal output on the Phi safety corpus (Appendix~\ref{app:examples}).
We treat these multiclass transitions as exploratory and make quantitative claims from the binary refusal endpoint.
Across Qwen's 2,400 outputs, inter-judge agreement was 76.0\% unanimous and 97.2\% by majority (Appendix~\ref{app:judges}).

\FloatBarrier

\section{Cipher stress test}
\label{sec:cipher_case}

\subsection{Question, tested cipher, and endpoint}
\label{sec:bypass}

We test whether gate effects depend on a gate-readable detection-layer representation.
A Latin substitution cipher taught in context changes the input form while leaving the model to infer the mapping from examples.
We use it as a bounded case study, not as a comprehensive encoding benchmark.

Cipher interchange differs from the discovery and scaling interchange in Section~\ref{sec:qwen_interchange}.
Instead of scoring a teacher-forced continuation by NLL, it measures total attribution along a frozen residual-space sensitive-minus-control probe direction.
We report mean absolute pairwise necessity and sufficiency because signed effects cancel across these heterogeneous corpora (Appendix~\ref{app:methods}).

\subsection{Weakened gate-readable signal and selected-head effect}
\label{sec:cipher_gate}

Cipher encoding sharply reduces the measured detection signal across three models from three labs.
In Qwen3-8B ($n{=}120$), the peak-layer score falls 65.9\% from plaintext under cipher (142.3$\to$48.5 at L35) and lies 56.1\% below the benign mean (110.5).
The safety-domain detection signal drops 88\% in Phi-4-mini ($n{=}120$, 37.1$\to$4.3 at L16) and 70\% in Gemma-2-2B ($n{=}120$, 97.6$\to$28.9 at L14).
In the automated behavior analyses, cipher-encoded prompts predominantly elicit decoding or puzzle-solving responses rather than refusal.

Under the tested cipher, mean-absolute necessity decreases by approximately 99\% in Phi-4-mini and Gemma-2-2B and by 70\% in Qwen3-8B.
Mean-absolute sufficiency decreases by approximately 86\% in Phi and Gemma and 35\% in Qwen.
Under this endpoint, the selected gate activation has much less effect under cipher, although Qwen retains a measurable residual.
These values measure a causal intervention on routing-direction attribution; generated refusal is a separate endpoint.
Layer-by-layer probe scores show a temporal separation: at the gate layer, cipher prompts track benign; at deeper layers (L24--29 in Phi-4), the probe score rises above benign after the measured gate layer, consistent with being unavailable to that gate at its intervention point.

Appendix Figure~\ref{fig:cipher_interchange} reports the three-model comparison.

\subsection{Rescue bounds}
\label{sec:cipher_rescue}

The automated heuristic labels 58/120 Phi-4-mini outputs as refusals after injection of the matched plaintext gate activation, compared with 0/120 under cipher alone; no human validation is available for this endpoint.
For Qwen3-8B, a separate $n{=}120$ test covers the gate alone and nested 3-, 6-, and 10-head attention groups.
The same conservative automated diagnostic finds 0/120 refusals after every injection, with only 9/120 plaintext diagnostic refusals.
Given the final-prompt-token, attention-only intervention and sparse plaintext baseline, the Qwen result is a narrow automated null.
It cannot adjudicate a distributed rescue mechanism or determine whether deeper layers reconstruct the encoded request.
Appendix~\ref{app:bijection} gives the full protocols and disclosures.

\subsection{Cipher contrast as a complementary screen}
\label{sec:cipher_contrast}

The cipher condition also permits a low-cost screen for heads whose direct probe-direction attribution changes with input form.
For each head, the \emph{cipher contrast score} is the absolute difference between its mean plaintext and cipher attribution under a frozen sensitive-minus-control probe direction.
The screen recovers known gate and amplifier loci near the top of the rankings while nominating additional candidates that were not interchanged.
Because exact rankings vary with the direction and statistic, we use this method as coarse group-enrichment evidence rather than a direction-independent circuit map.
Appendix~\ref{app:cipher_diagnostic} gives the thresholds, rankings, sensitivity analysis, and full scatterplot.

\subsection{Interpretation}
\label{sec:cipher_interpretation}

Under the tested substitution cipher, inputs fail to produce the measured gate-readable routing signal at the gate layer, placing the observed failure at a particular time and interface.
Failed semantic binding and correlates of formal cipher processing both fit the evidence; deeper semantic reconstruction remains unresolved.
We limit the claim to this cipher.

\FloatBarrier

\section{Discussion}
\label{sec:discussion}

\subsection{Synthesis}
\label{sec:synthesis}

The experiments separate three stages that behavior alone conflates.
Policy-relevant content is linearly decodable before generation, but detection does not determine the response.
In Qwen3-8B, output attribution identifies deep carriers while missing L17.H17, the strongest joint entry-trigger candidate among the ten heads tested by interchange.
Knockout then shows that five of six tested later-head contributions depend on that candidate.
The result is a localized trigger within a distributed cascade, not a discrete head that contains or executes the policy by itself.

Phi-4-mini and Gemma-2-2B reproduce this functional ordering at different locations, with partial dependence among later heads.
The nine additional models broaden the cross-model screen but do not supply the same four-method evidence.
Generated behavior also respects the distinction between local and distributed control: a cross-layer Phi intervention produces a large human-scored refusal reduction, while its single-head and within-layer estimates are inconclusive and the Gemma band endpoint is negative.
The cipher study places the tested failure at a time-sensitive routing interface by combining reduced gate-readable signal, weakened selected-gate effects, and bounded rescue tests.

\subsection{Relation to prior work}
\label{sec:related}

\paragraph{Representations of harmfulness and refusal.}
Refusal can be mediated by low-dimensional residual-stream structure \citep{arditi2024refusal,wollschlager2025geometry}, and representation-level interventions can alter refusal or censorship behavior \citep{zou2023representation,cyberey2025steering,garciaferrero2025refusal}.
Most directly, \citet{zhao2025harmfulness} establish that harmfulness and refusal occupy distinct representational directions, intervene on those directions, and use the retained harmfulness representation as a guard.
Their result motivates our component-level question: once policy-relevant content is represented, which components help map it into a policy response?
We localize one intervenable circuit interface in that mapping, not where a general refusal direction originates.

\paragraph{Safety heads and circuit localization.}
\citet{zhou2025safetyheads} attribute safety behavior to sparse attention heads, while \citet{chu2025keyheads} distinguish detection- and refusal-head candidates and intervene on them.
Concurrent work studies attention heads that are suppressed or remain active under jailbreaks \citep{yin2026robustheads}.
Our analysis tests directed dependence: activation interchange identifies upstream trigger candidates, and knockout cascades test whether those candidates influence later-head contributions.
We perform this four-method analysis in three architectures and treat the other nine models as screening analyses.
Methodologically, this builds on interchange interventions and activation patching \citep{geiger2021causal,meng2022locating,wang2023interpretability}.

\paragraph{Jailbreak robustness and encoded inputs.}
HarmBench and JailbreakBench standardize behavioral evaluation across attacks, models, and defenses \citep{mazeika2024harmbench,chao2024jailbreakbench}.
The closest antecedent to our cipher experiment is the bijection-learning attack of \citet{huang2025bijection}, which teaches randomly generated encodings in context and measures their behavioral effectiveness.
Substitution ciphers have also been used to study in-context task learning and internal decoding representations \citep{fang2025iclciphers}.
Our cipher study neither introduces this attack class nor provides a comprehensive robustness benchmark.
It uses a restricted substitution-cipher setting as a causal case study of when a known behavioral failure enters the routing circuit.

\paragraph{Political censorship.}
Prior work studies censorship behavior, elicitation, and forbidden-topic discovery \citep{casademunt2026censored,pan2026political,rager2025discovering}.
We use this setting as one testbed for component-level causal localization.

\subsection{Scope, limitations, and dual use}
\label{sec:limitations}
\label{sec:dual_use}

The four-method analysis is limited to Qwen3-8B, Phi-4-mini, and Gemma-2-2B.
The nine broader screens do not prove head-level homology, and behavior beyond 72B parameters remains unknown.
Attribution, teacher-forced NLL, and probe-direction changes are internal endpoints rather than substitutes for generation.
The strongest human-scored intervention evidence is the cross-layer Phi result; behavioral effects are not uniform across architectures or intervention groups.

The empirical domains are Chinese-political censorship and harmful-request refusal.
Sycophancy, privacy protection, bias mitigation, tool-use constraints, and policies without refusal-shaped endpoints may be organized differently.
Only binary refusal has been calibrated quantitatively with human labels; finer output categories remain machine-assigned and exploratory.
The MLP pathway has not been decomposed at the feature level, and the current instrumentation does not cover every architecture or interface.

The cipher analysis primarily tests one Latin substitution setup across three models.
Smaller Qwen tests with Base64 and Chinese-character substitutions are exploratory rather than an encoding-family evaluation.
Later probe-direction signal does not establish semantic reconstruction, so the analysis remains agnostic between failed semantic binding and correlates of formal processing.

The encoding attack class and in-context bijection-learning protocol predate this work, but mechanistic localization can still lower the effort needed to analyze or adapt known bypasses.
White-box head interventions may likewise be misused against open-weight models.
Our threat model therefore includes black-box users applying previously published encoding attacks and white-box users with access to model activations.

We retain the detail needed to reproduce the causal claims because the same measurements support defensive auditing.
We remove redundant worked mappings from the main text and do not release model weights, modified checkpoints, or a deployment-facing attack service.
The released code instruments public open-weight models for research.
The results suggest concrete defensive tests: evaluate policies under meaning-preserving encodings, avoid relying on a single early representation, and test whether deeper or redundant detectors can recover missed signals.
These are design implications, not defenses validated by this paper.

\subsection{Conclusion}
\label{sec:conclusion}

Policy routing is sparse without being pointlike: a localized trigger recruits a distributed, architecture-specific cascade.
This explains why output DLA can miss a causal gate, why local and cross-layer interventions differ, and why altered input form can disrupt routing before related signal appears downstream.

Audits should pair behavioral benchmarks with interventions that separate detection, routing, and output, avoid assuming stable head locations across generations, and test altered input forms rather than rely on one early representation.

\begingroup
\setlength{\bibsep}{5pt plus 1pt minus 1pt}
\small
\bibliographystyle{plainnat}
\bibliography{How_Alignment_Routes_references}

@article{frank2026detection,
  author = {Frank, Gregory N.},
  title = {Detection Is Cheap, Routing Is Learned: Why Refusal-Based Alignment Evaluation Fails},
  journal = {arXiv preprint arXiv:2603.18280},
  year = {2026},
  eprint = {2603.18280},
  archivePrefix = {arXiv},
  url = {https://arxiv.org/abs/2603.18280}
}

@article{arditi2024refusal,
  author = {Arditi, Andy and Obeso, Oscar and Syed, Aaquib and Paleka, Daniel and Panickssery, Nina and Gurnee, Wes and Nanda, Neel},
  title = {Refusal in Language Models Is Mediated by a Single Direction},
  journal = {arXiv preprint arXiv:2406.11717},
  year = {2024},
  eprint = {2406.11717},
  archivePrefix = {arXiv},
  url = {https://arxiv.org/abs/2406.11717}
}

@article{casademunt2026censored,
  author = {Casademunt, Helena and Cywi{\'n}ski, Bartosz and Tran, Khoi and Jakkli, Arya and Marks, Samuel and Nanda, Neel},
  title = {Censored {LLM}s as a Natural Testbed for Secret Knowledge Elicitation},
  journal = {arXiv preprint arXiv:2603.05494},
  year = {2026},
  eprint = {2603.05494},
  archivePrefix = {arXiv},
  url = {https://arxiv.org/abs/2603.05494}
}

@article{cyberey2025steering,
  author = {Cyberey, Hannah and Evans, David},
  title = {Steering the {CensorShip}: Uncovering Representation Vectors for {LLM} ``Thought'' Control},
  journal = {arXiv preprint arXiv:2504.17130},
  year = {2025},
  eprint = {2504.17130},
  archivePrefix = {arXiv},
  url = {https://arxiv.org/abs/2504.17130}
}

@article{garciaferrero2025refusal,
  author = {Garc{\'i}a-Ferrero, Iker and Montero, David and Orus, Roman},
  title = {Refusal Steering: Fine-grained Control over {LLM} Refusal Behaviour for Sensitive Topics},
  journal = {arXiv preprint arXiv:2512.16602},
  year = {2025},
  eprint = {2512.16602},
  archivePrefix = {arXiv},
  url = {https://arxiv.org/abs/2512.16602}
}

@inproceedings{hewitt2019designing,
  author = {Hewitt, John and Liang, Percy},
  title = {Designing and Interpreting Probes with Control Tasks},
  booktitle = {Proceedings of the 2019 Conference on Empirical Methods in Natural Language Processing and the 9th International Joint Conference on Natural Language Processing ({EMNLP-IJCNLP})},
  year = {2019},
  pages = {2733--2743},
  doi = {10.18653/v1/D19-1275},
  url = {https://doi.org/10.18653/v1/D19-1275}
}

@article{pan2026political,
  author = {Pan, Jennifer and Xu, Xu},
  title = {Political censorship in large language models originating from {China}},
  journal = {PNAS Nexus},
  year = {2026},
  volume = {5},
  number = {2},
  pages = {pgag013},
  doi = {10.1093/pnasnexus/pgag013},
  url = {https://doi.org/10.1093/pnasnexus/pgag013}
}

@article{wollschlager2025geometry,
  author = {Wollschl{\"a}ger, Tom and Elstner, Jannes and Geisler, Simon and Cohen-Addad, Vincent and G{\"u}nnemann, Stephan and Gasteiger, Johannes},
  title = {The Geometry of Refusal in Large Language Models: Concept Cones and Representational Independence},
  journal = {arXiv preprint arXiv:2502.17420},
  year = {2025},
  eprint = {2502.17420},
  archivePrefix = {arXiv},
  url = {https://arxiv.org/abs/2502.17420}
}

@inproceedings{zhao2025harmfulness,
  author = {Zhao, Jiachen and Huang, Jing and Wu, Zhengxuan and Bau, David and Shi, Weiyan},
  title = {{LLM}s Encode Harmfulness and Refusal Separately},
  booktitle = {Advances in Neural Information Processing Systems},
  volume = {38},
  year = {2025},
  url = {https://proceedings.neurips.cc/paper_files/paper/2025/hash/cd18539787d90e1d682d557c2c71b534-Abstract-Conference.html}
}

@article{zou2023representation,
  author = {Zou, Andy and Phan, Long and Chen, Sarah and Campbell, James and Guo, Phillip and Ren, Richard and Pan, Alexander and Yin, Xuwang and Mazeika, Mantas and Dombrowski, Ann-Kathrin and Goel, Shashwat and Li, Nathaniel and Byun, Michael J. and Wang, Zifan and Mallen, Alex and Basart, Steven and Koyejo, Sanmi and Song, Dawn and Fredrikson, Matt and Kolter, J. Zico and Hendrycks, Dan},
  title = {Representation Engineering: A Top-Down Approach to {AI} Transparency},
  journal = {arXiv preprint arXiv:2310.01405},
  year = {2023},
  eprint = {2310.01405},
  archivePrefix = {arXiv},
  url = {https://arxiv.org/abs/2310.01405}
}

@inproceedings{geiger2021causal,
  title={Causal Abstractions of Neural Networks},
  author={Geiger, Atticus and Lu, Hanson and Icard, Thomas and Potts, Christopher},
  booktitle={Advances in Neural Information Processing Systems (NeurIPS)},
  year={2021}
}

@article{meng2022locating,
  title={Locating and Editing Factual Associations in {GPT}},
  author={Meng, Kevin and Bau, David and Andonian, Alex and Belinkov, Yonatan},
  journal={Advances in Neural Information Processing Systems (NeurIPS)},
  year={2022}
}

@inproceedings{wang2023interpretability,
  title={Interpretability in the Wild: a Circuit for Indirect Object Identification in {GPT-2} Small},
  author={Wang, Kevin and Variengien, Alexandre and Conmy, Arthur and Shlegeris, Buck and Steinhardt, Jacob},
  booktitle={International Conference on Learning Representations (ICLR)},
  year={2023}
}

@article{rager2025discovering,
  title={Discovering Forbidden Topics in Language Models},
  author={Rager, Can and Wendler, Chris and Gandikota, Rohit and Bau, David},
  journal={arXiv preprint arXiv:2505.17441},
  year={2025}
}

@inproceedings{zhou2025safetyheads,
  author = {Zhou, Zhenhong and Yu, Haiyang and Zhang, Xinghua and Xu, Rongwu and Huang, Fei and Wang, Kun and Liu, Yang and Fang, Junfeng and Li, Yongbin},
  title = {On the Role of Attention Heads in Large Language Model Safety},
  booktitle = {The Thirteenth International Conference on Learning Representations},
  year = {2025},
  url = {https://openreview.net/forum?id=h0Ak8A5yqw}
}

@inproceedings{chu2025keyheads,
  author = {Chu, Kuan-Lin and Sun, Chung-En and Weng, Tsui-Wei},
  title = {How to Make {LLM}s Safer? Detecting and Editing Key Heads in {LLM}s},
  booktitle = {NeurIPS 2025 Workshop on Lock-{LLM}},
  year = {2025},
  url = {https://openreview.net/forum?id=SNwwxvM0cx}
}

@inproceedings{huang2025bijection,
  author = {Huang, Brian R. Y. and Li, Maximilian and Tang, Leonard},
  title = {Endless Jailbreaks with Bijection Learning},
  booktitle = {The Thirteenth International Conference on Learning Representations},
  year = {2025},
  url = {https://openreview.net/forum?id=xP1radUi32}
}

@inproceedings{mazeika2024harmbench,
  author = {Mazeika, Mantas and Phan, Long and Yin, Xuwang and Zou, Andy and Wang, Zifan and Mu, Norman and Sakhaee, Elham and Li, Nathaniel and Basart, Steven and Li, Bo and Forsyth, David and Hendrycks, Dan},
  title = {{H}arm{B}ench: A Standardized Evaluation Framework for Automated Red Teaming and Robust Refusal},
  booktitle = {Proceedings of the 41st International Conference on Machine Learning},
  series = {Proceedings of Machine Learning Research},
  volume = {235},
  pages = {35181--35224},
  year = {2024},
  publisher = {PMLR},
  url = {https://proceedings.mlr.press/v235/mazeika24a.html}
}

@inproceedings{chao2024jailbreakbench,
  author = {Chao, Patrick and Debenedetti, Edoardo and Robey, Alexander and Andriushchenko, Maksym and Croce, Francesco and Sehwag, Vikash and Dobriban, Edgar and Flammarion, Nicolas and Pappas, George J. and Tram{\`{e}}r, Florian and Hassani, Hamed and Wong, Eric},
  title = {{J}ailbreak{B}ench: An Open Robustness Benchmark for Jailbreaking Large Language Models},
  booktitle = {Advances in Neural Information Processing Systems},
  volume = {37},
  year = {2024},
  url = {https://proceedings.neurips.cc/paper_files/paper/2024/hash/63092d79154adebd7305dfd498cbff70-Abstract-Datasets_and_Benchmarks_Track.html}
}

@article{fang2025iclciphers,
  author = {Fang, Zhouxiang and Mishra, Aayush and Gao, Muhan and Liu, Anqi and Khashabi, Daniel},
  title = {{ICL CIPHERS}: Quantifying ``Learning'' in In-Context Learning via Substitution Ciphers},
  journal = {arXiv preprint arXiv:2504.19395},
  year = {2025},
  eprint = {2504.19395},
  archivePrefix = {arXiv},
  url = {https://arxiv.org/abs/2504.19395}
}

@article{yin2026robustheads,
  author = {Yin, Yanchen and Han, Dongqi and Li, Linghui},
  title = {Robust Harmful Features Under Jailbreak Attacks: Mechanistic Evidence from Attention Head Specialization in Large Language Models},
  journal = {arXiv preprint arXiv:2606.28153},
  year = {2026},
  eprint = {2606.28153},
  archivePrefix = {arXiv},
  note = {Accepted at ICML 2026},
  url = {https://arxiv.org/abs/2606.28153}
}
\endgroup

\begin{ack}
The author thanks the condition-blinded annotators who volunteered their time for the behavioral evaluation.

\textbf{Funding.}
The author conducted and self-funded this work before joining MACE AI (formerly Molt AI Corp.).
No third party provided financial support.
Volunteer annotators provided uncompensated in-kind support for the behavioral evaluation.

\textbf{Competing interests.}
The author is currently employed by MACE AI, as stated in the affiliation, and declares no other competing financial interests.
\end{ack}

\appendix
\clearpage

\section{Mechanistic methods}
\label{app:methods}

\paragraph{Output logit-DLA.}
For matched pair~$i$, let $R$ be a fixed set of refusal-token IDs and let $a_i$ be the first meaningful token in a short greedy continuation from the control prompt.
Writing $u_t=W_U[t]$ for the unembedding vector of token~$t$, we define
\[
d_i^{\mathrm{logit}}
=
\frac{1}{|R|}\sum_{r\in R}u_r-u_{a_i}.
\]
The same pair-specific direction is used for the sensitive and control conditions.

Let $x_{ic}^{(p)}$ be component~$c$'s residual-stream contribution at the last prompt token under condition $p\in\{s_i,c_i\}$, and let $r_i^{(p)}$ be that condition's full residual at the same position.
We use a fixed-scale linearization of the final normalization, evaluating its scale on $r_i^{(p)}$ and holding that scale fixed for each component:
\[
\Lambda_r(x)=
\begin{cases}
\displaystyle
\gamma\odot x\Big/\sqrt{D^{-1}\lVert r\rVert_2^2+\epsilon},
& \text{RMSNorm},\\[6pt]
\displaystyle
\gamma\odot\bigl(x-\mu(x)\mathbf{1}\bigr)
\Big/\sqrt{D^{-1}\lVert r-\mu(r)\mathbf{1}\rVert_2^2+\epsilon},
& \text{LayerNorm},
\end{cases}
\]
where $D$ is the residual width and $\gamma$ is the learned norm weight.
This is not equivalent to applying the nonlinear final norm independently to each component.

The component attribution is
\[
A_{ic}^{(p)}
=
\left\langle \Lambda_{r_i^{(p)}}\!\left(x_{ic}^{(p)}\right),
d_i^{\mathrm{logit}}\right\rangle,
\qquad
\Delta\mathrm{DLA}_{ic}
=
A_{ic}^{(s_i)}-A_{ic}^{(c_i)}.
\]
Positive $\Delta\mathrm{DLA}$ favors refusal relative to the matched control answer; negative values indicate counter-routing.
The original head screen ranks $\left|n^{-1}\sum_i\Delta\mathrm{DLA}_{ih}\right|$.
This quantity is a direct output attribution under the stated linearization, not a causal effect.
For attention head~$h$, the residual contribution is $x_{ih}=W_{O,h}z_{ih}$, where $z_{ih}$ is the head's pre-output-projection activation and $W_{O,h}$ is the corresponding slice of the attention output matrix.

\paragraph{Directional head intervention.}
This intervention uses a frozen residual-space detection direction~$v$, not the output-logit direction above.
For candidate head~$h$, we solve through its output-projection slice and normalize the resulting head-space direction:
\[
q_h
=
\operatorname{normalize}
\left(
\arg\min_q \lVert W_{O,h}q-v\rVert_2^2
\right).
\]
At every sequence position, the live head activation is transformed as
\[
z'_{h,t}
=
z_{h,t}-\alpha\langle z_{h,t},q_h\rangle q_h,
\]
with $\alpha=5$ in the head-localization and model-panel runs.
Because $\alpha\ne1$, we call this a directional intervention rather than an exact orthogonal projection.

For each matched pair, we greedily generate a short continuation $y_i$ from the control prompt.
We then score its mean per-token teacher-forced negative log-likelihood,
\[
\ell_i(p;\mathcal I)
=
-\frac{1}{|y_i|}
\sum_j
\log P\!\left(y_{ij}\mid p,y_{i,<j};\mathcal I\right),
\]
in nats/token.
The directional-intervention effect for head~$h$ is $B_{ih}=\ell_i(s_i)-\ell_i(s_i;\mathcal I_h)$.
A positive value means the intervention makes the matched-control continuation more likely under the sensitive prompt.
This NLL change, not DLA, is the endpoint reported for head-level intervention.

\paragraph{Activation interchange.}
Discovery and scaling interchange use the same matched-control-continuation NLL endpoint.
Let $\ell_i(p)$ denote its NLL under prompt~$p$ and let $G_i=\ell_i(s_i)-\ell_i(c_i)$ be the baseline gap.
For head~$h$, necessity is
$N_{ih}=\ell_i(s_i)-\ell_i(s_i;h\leftarrow c_i)$,
and sufficiency is
$S_{ih}=\ell_i(c_i;h\leftarrow s_i)-\ell_i(c_i)$.
Positive necessity means the control activation makes the control continuation more likely under the sensitive prompt; positive sufficiency means the sensitive activation makes it less likely under the control prompt.
Reported percentages normalize $\operatorname{mean}_i N_{ih}$ and $\operatorname{mean}_i S_{ih}$ by $\operatorname{mean}_i G_i$.
The implementation caches the head's complete prompt-length activation sequence and replaces its slice at every overlapping prompt-token position; continuation positions remain live during teacher-forced scoring.
Once candidates are fixed, no logit-difference direction enters this interchange endpoint.
A head with joint necessity and sufficiency is an upstream trigger candidate under this endpoint; heads with necessity but little sufficiency are downstream amplifier candidates.
For Qwen3-8B, the ten-head set comprised the highest-ranked directional-intervention candidates on the discovery corpus; the directional stage followed the all-head output-DLA screen, and interchange did not search all 1,152 heads.

\begin{figure}[tbp]
\centering
\includegraphics[width=\textwidth]{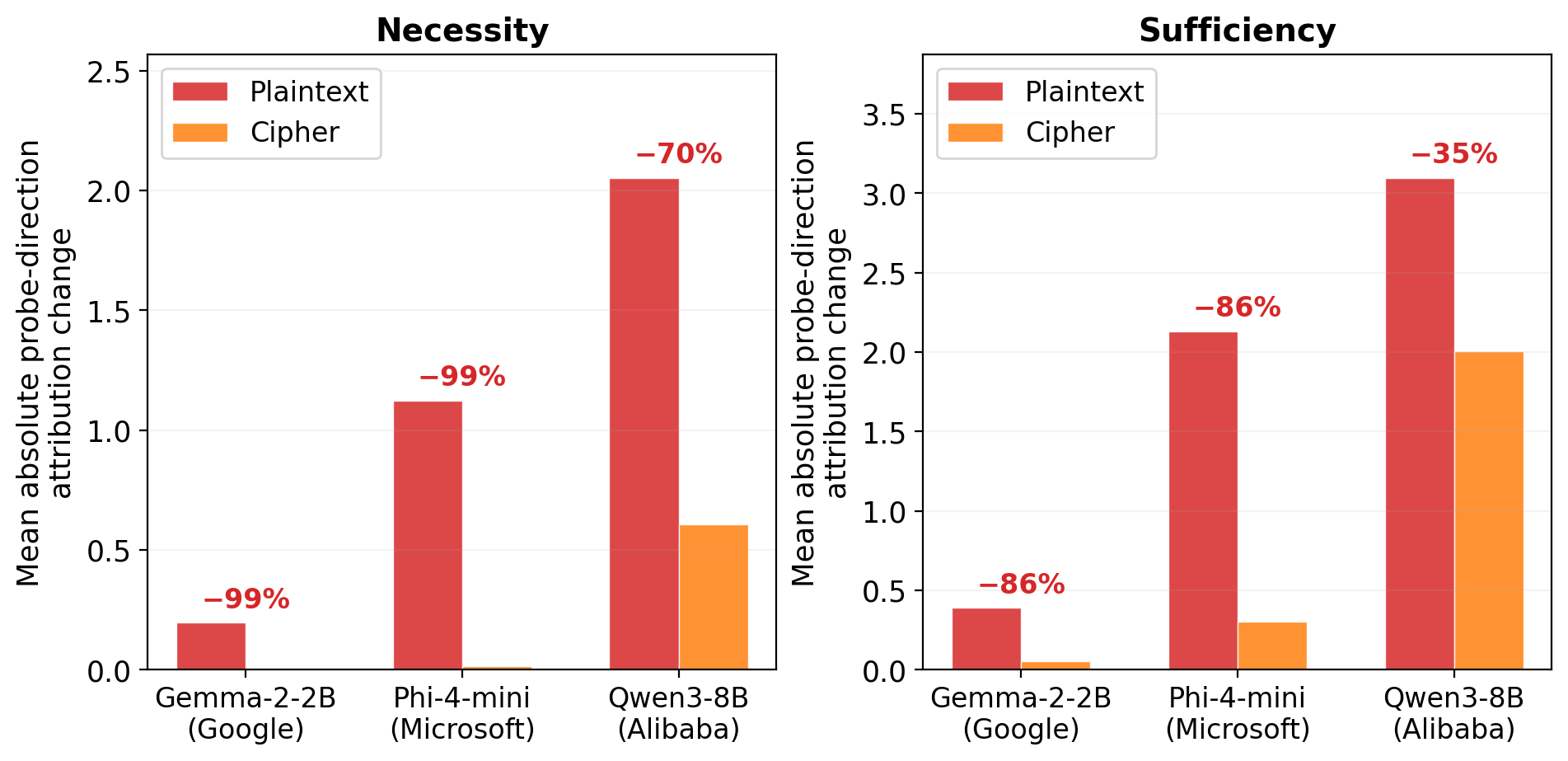}
\caption{%
\textbf{Selected-gate interchange effects on routing-direction attribution under the tested cipher ($n{=}120$).}
Bars report mean absolute pairwise necessity and sufficiency after projection onto each model's frozen sensitive-minus-control probe direction.
The effect decreases sharply in all three models, with a larger residual effect in Qwen.
This endpoint is distinct from both output logit-DLA and the continuation-NLL interchange used for discovery and scaling.%
}
\label{fig:cipher_interchange}
\end{figure}

\paragraph{Cipher-interchange endpoint.}
The cipher experiment uses activation interchange but scores a different internal endpoint.
For prompt~$p$, let
\[
A_v(p)
=
\sum_c
\left\langle
\Lambda_{r(p)}\!\left(x_c(p)\right),v
\right\rangle,
\]
where $v$ is a frozen residual-space sensitive-minus-control probe direction and the sum covers direct attention-head and MLP contributions at the last prompt token.
Cipher-interchange necessity and sufficiency are the pairwise changes in $A_v$ after swapping the gate activation with the corresponding matched activation.
Figure~\ref{fig:cipher_interchange} reports the mean absolute pairwise effects, $n^{-1}\sum_i|N_{ih}^{(v)}|$ and $n^{-1}\sum_i|S_{ih}^{(v)}|$, because signed effects cancel across heterogeneous prompts.
This endpoint is probe-direction attribution, not output logit-DLA or continuation NLL.

\paragraph{Knockout cascade.}
We zero the gate head's o\_proj input slice (all $d_h$ dimensions) via a forward pre-hook, effectively removing that head's contribution from all subsequent computation.
We then re-run the ordinary output logit-DLA decomposition and measure each selected head's knockout-minus-normal change.
Results are averaged over the full prompt corpus ($n{=}120$ in Qwen, Phi, and Gemma), with per-pair raw data retained for bootstrap validation.
An exploratory specificity control on 20 pairs compared the Qwen gate with 10 random non-gate heads at similar depths.
Mean cascade suppression was 10.51\% for the gate and 3.91\% for the random heads (SD 2.09\%; maximum 7.66\%).
The prespecified Gemma comparison set contained two pre-gate DLA candidates, L12.H3 and L8.H3.
Their reported output-normalized DLA changes can reflect the altered final-residual normalization and are not counted as downstream cascade evidence.

\paragraph{Prefix-restricted probe-direction attribution.}
For the intermediate analysis, we project the sensitive-minus-control difference in each head's output onto the frozen residual-space probe direction and rank only heads written before a given layer cutoff.
This is not output-logit DLA and does not measure a fixed head's signal propagating through successive residual checkpoints.
At the L18 cutoff, Qwen L17.H17 has the second-largest absolute mean probe-direction contribution among head outputs written through L17.
Its value remains fixed as later heads enter the ranking, so causal claims about downstream dependence come from the knockout cascade rather than rank dilution in this diagnostic.

\paragraph{Parallel MLP-pathway diagnostics.}
The direct MLP share is corpus-dependent: approximately 23\% on the diverse $n{=}120$ corpus and approximately 61\% on concentrated single-topic prompts.
We test this pathway in Qwen3-8B ($n{=}120$) using activation interchange and knockout scored by direct attribution to the same frozen residual-space probe direction.
The top MLP layers have mean-absolute interchange necessity of 5.2--8.7 and knockout effects up to approximately 70, larger than the attention gate under this measure.
Using \texttt{mlp\_share} $\geq$ the sample median to define the high-share group gives 61 high-share and 59 low-share prompts; L17.H17's mean-absolute necessity is 1.84 and 2.22 in the two groups, respectively.
This diagnostic shows that a larger direct MLP share does not eliminate the gate effect on the internal probe-direction endpoint; it does not establish generated-output necessity or identify the relevant MLP features.

\paragraph{Direction robustness.}
The logit-DLA ranking depends on its refusal set and answer target.
At $n{=}120$, an audit using four closely related definitions finds high all-head agreement among the prompt-conditioned variants (Spearman $\rho=0.994$--$0.997$; identical top-five sets).
A fixed ``The'' answer target retains high overall correlation ($\rho=0.982$--$0.986$) but changes the extreme set (top-five Jaccard $=0.25$).
L17.H17 remains a low-output-attribution head under all four definitions, at ranks 277--314.
Cipher-contrast rankings are less stable across these definitions ($\rho=0.705$--$0.896$; top-10 Jaccard $=0.429$--$0.667$).
Relative to the original residual-space probe-direction map, the correlations are $\rho=0.638$--$0.678$.
Five or six of the seven frozen causal heads nevertheless remain in the top quartile under every output-logit definition, supporting group enrichment but not an invariant exact ranking.
This audit uses the control prompt's top next-token target and ranks mean absolute pairwise differences, whereas the original screen uses the first meaningful greedy control token and ranks the absolute signed mean; it is therefore a sensitivity analysis, not a literal rerun with only one target changed.
The fixed-candidate discovery/scaling interchange result is unaffected because its endpoint is continuation NLL rather than a logit direction.
Bootstrap resampling assesses sampling stability, not robustness to measurement definition.

\paragraph{Candidate-selection sensitivity.}
For Llama-3.3-70B, the DLA screen nominated L77.H47, with interchange necessity of 1.3\%.
A separate cipher-contrast screen nominated L26.H40, whose interchange necessity was 2.0\% but whose mean sufficiency was slightly negative.
We therefore treat L26.H40 as a higher-necessity routing candidate rather than a trigger signature.
The comparison illustrates that candidate discovery and the exact selected locus depend on the screen.

\paragraph{Detection-layer modulation.}
We add or subtract $\alpha \cdot d$ at the detection layer via a forward hook.
Alpha sweeps run from 0 to 50 in increments of~5, with both attenuation ($-\alpha$) and amplification ($+\alpha$).

\paragraph{Statistical validation.}
For ranking stability, we resample matched prompt pairs with replacement 2,000 times and compare each resampled top-$K$ set with the full-sample set by Jaccard overlap.
For Qwen interchange, the joint statistic is $T_h=[\overline{N}_h]_+[\overline{S}_h]_+$.
The paired sign-flip null applies one random sign jointly to the necessity and sufficiency observations for each prompt pair, preserving their within-pair dependence and dependence across tested heads.
We use 10,000 draws and evaluate both L17.H17 and the maximum $T_h$ across the ten tested candidates.
Monte Carlo values use the add-one convention $p=(b+1)/(B+1)$, where $b$ is the number of draws that match or exceed the observed statistic and $B=10{,}000$.
When $b=0$, we report $p\approx10^{-4}$ rather than $p=0$ or $p<10^{-4}$; Table~\ref{tab:interchange_null} gives both comparisons.
For the knockout cascades, we estimate uncertainty separately for each downstream head using 20,000 paired bootstrap resamples of $\Delta\mathrm{DLA}^{\mathrm{knockout}}-\Delta\mathrm{DLA}^{\mathrm{normal}}$.
The resulting 95\% percentile intervals are pointwise; negative values indicate suppression.

\begin{table}[H]
\caption{Paired sign-flip calibration of the Qwen L17.H17 joint interchange statistic ($n{=}120$; 10,000 draws). Monte Carlo $p$-values use the add-one convention.}
\label{tab:interchange_null}
\centering
\small
\begin{tabular}{@{}lrrrrr@{}}
\toprule
Comparison & Observed & Null mean & Ratio & Exceedances & Add-one $p$ \\
\midrule
Headwise L17.H17 & $9.18\times10^{-4}$ & $1.32\times10^{-5}$ & 69.6 & 0/10,000 & $1.00\times10^{-4}$ \\
Maximum across 10 heads & $9.18\times10^{-4}$ & $2.22\times10^{-5}$ & 41.4 & 0/10,000 & $1.00\times10^{-4}$ \\
\bottomrule
\end{tabular}
\end{table}

\paragraph{Behavioral classification.}
Three independent LLM judges (Gemini 2.0 Flash, Llama 3.1 8B, GPT-4o-mini) classify dose-response outputs into six categories (REFUSAL, FACTUAL, STEERED, HARMFUL\_GUIDANCE, INCOHERENT, EVASION) at temperature~0.
Final label: majority vote; three-way disagreements labeled DISAGREE.
Agreement: 76.0\% unanimous on Qwen ($n{=}2{,}400$); 84.0\% on Phi-4-mini ($n{=}2{,}400$).
Disagreement concentrates on adjacent categories: REFUSAL dissenters label EVASION (17\%); FACTUAL dissenters label EVASION or STEERED (15\%); STEERED is the least reliable (45\% unanimous).
REFUSAL and FACTUAL, the categories that anchor the dose-response curves, have 78\% and 83\% unanimity respectively.

\section{Generated-behavior validation}
\label{app:behavioral_validation}

\paragraph{Phi head-group interventions.}
We generated 720 outputs from 120 prompts under six prespecified conditions: baseline, L13.H7 alone, a five-head L13 group, a seven-head L16 group, a cross-layer top-10 group, and a legacy mixed 26-head group.
The analysis used a balanced 198-response subset, 33 per condition.
Two condition-blinded raters independently assigned binary refusal labels and agreed on 194/198 responses (Cohen's $\kappa=0.948$); a third condition-blinded rater adjudicated the four disagreements.

The reduced double-coding design was adopted after the first rater attempted 320 of the 720 randomized positions.
Subset selection used packet position, prompt-pair availability, and intervention condition without reference to the response text, human labels, automated scores, routing measurements, or condition effects.
Aggregate first-rater counts had been seen during file quality assurance.
The analysis is therefore a post hoc, value-blind validation subset rather than a preregistered behavioral endpoint.

The main comparison uses unadjusted Newcombe hybrid-score intervals for differences in proportions and Holm-adjusted Fisher exact tests for the five contrasts with baseline.
Because the balanced subset was not fully paired, we also report exact same-prompt sensitivity analyses.
For the cross-layer top-10 group, 11 of 15 shared prompts changed from refusal to non-refusal and none reversed (Holm-adjusted exact McNemar $p=0.00391$).
For the legacy group, the corresponding count was 15 of 25, again with no reversals ($p=0.000305$).

\paragraph{Internal measure versus behavior.}
For 96 intervention cells with a human-labeled baseline for the same prompt, we compared refusal with the change in a frozen total residual-stream projection onto a mid-layer sensitive-minus-control mean-difference direction.
This quantity is not per-head logit-DLA or interchange.
Pooling conditions gives Pearson $r=0.43$ with a prompt-clustered 95\% interval of $[0.25,0.61]$; Spearman $\rho=0.44$.
After removing condition-level differences, Pearson $r=-0.09$ (95\% CI $[-0.38,0.22]$; condition-stratified permutation $p=0.475$).
Three of five intervention conditions have almost no within-condition outcome variation, so the adjusted estimate is weakly constrained.
We do not treat this internal measure as a prompt-level behavioral surrogate.

\paragraph{Binary-refusal calibration for dose response.}
Two condition-blinded raters agreed on 107/120 binary labels (Cohen's $\kappa=0.779$), and a third condition-blinded rater adjudicated the 13 disagreements.
Among machine-majority strata covered by the audit, post-stratified accuracy of the automated binary mapping was 96.8\% for Qwen (2,331/2,400 outputs covered) and 88.3\% for Phi (2,349/2,400 covered); unweighted accuracy in the category-balanced audit sample was 83.3\% and 66.0\%, respectively.
Within those covered strata, human calibration raises the estimated aggregate refusal rate from 4.7\% to 7.9\% for Qwen and from 57.9\% to 69.7\% for Phi.
The audit does not provide a condition-specific correction factor and does not validate the six-way output taxonomy.

\paragraph{Gemma negative endpoint.}
The Gemma-2-2B generation screen used 30 prompts per condition and an automated heuristic.
Its all-band intervention changed refusal from 30/30 to 29/30.
We retain this negative result because internal gate-to-amplifier dependence does not imply a uniform generated-behavior effect across architectures or intervention groups.

\section{Knockout-cascade uncertainty}
\label{app:cascade_intervals}

Table~\ref{tab:cascade_intervals} reports pointwise intervals for every selected comparison head.
Each change is ordinary output logit-DLA after gate knockout minus the value in the unmodified run, so negative values indicate suppression.
The two Gemma rows that precede the L13 gate are retained for transparency but are not counted as downstream dependence.

\begin{table}[H]
\caption{Raw gate-knockout changes in selected-head output logit-DLA with pointwise 95\% paired-bootstrap intervals ($n{=}120$; 20,000 resamples per head). Gemma rows marked \emph{pre} precede the gate and are excluded from downstream counts.}
\label{tab:cascade_intervals}
\centering
\small
\begin{tabular}{@{}llrr@{}}
\toprule
Model & Comparison head & Knockout $-$ normal & 95\% CI \\
\midrule
Qwen3-8B & L22.H7 & $-0.028033$ & $[-0.036532,-0.019894]$ \\
Qwen3-8B & L23.H3 & $-0.002417$ & $[-0.004687,-0.000186]$ \\
Qwen3-8B & L22.H4 & $-0.003883$ & $[-0.007616,-0.000304]$ \\
Qwen3-8B & L22.H5 & $-0.005738$ & $[-0.009598,-0.001830]$ \\
Qwen3-8B & L22.H6 & $+0.018295$ & $[+0.013048,+0.023808]$ \\
Qwen3-8B & L23.H2 & $-0.019738$ & $[-0.027454,-0.012408]$ \\
\midrule
Phi-4-mini & L16.H13 & $+0.060232$ & $[+0.015807,+0.103545]$ \\
Phi-4-mini & L27.H19 & $-0.015464$ & $[-0.062728,+0.031650]$ \\
Phi-4-mini & L29.H18 & $-0.257382$ & $[-0.339258,-0.179512]$ \\
Phi-4-mini & L26.H11 & $-0.146517$ & $[-0.234916,-0.067368]$ \\
Phi-4-mini & L26.H9 & $-0.192096$ & $[-0.386175,-0.048561]$ \\
\midrule
Gemma-2-2B & L16.H1 & $-0.004885$ & $[-0.007862,-0.001948]$ \\
Gemma-2-2B & L15.H2 & $-0.009651$ & $[-0.011676,-0.007640]$ \\
Gemma-2-2B & L12.H3 (pre) & $-0.001025$ & $[-0.001292,-0.000763]$ \\
Gemma-2-2B & L8.H3 (pre) & $+0.000443$ & $[+0.000099,+0.000803]$ \\
Gemma-2-2B & L15.H7 & $+0.003337$ & $[+0.001879,+0.004838]$ \\
\bottomrule
\end{tabular}
\end{table}

\section{Evidence summary}
\label{app:evidence}

\begin{table}[tbp]
\caption{\textbf{Measurement roadmap.} The endpoints answer different questions and are not interchangeable.}
\label{tab:evidence}
\centering
\small
\begin{tabular}{@{}>{\raggedright\arraybackslash}p{3.0cm}>{\raggedright\arraybackslash}p{2.8cm}>{\raggedright\arraybackslash}p{3.2cm}>{\raggedright\arraybackslash}p{3.3cm}@{}}
\toprule
Question & Method & Endpoint & Supports \\
\midrule
Which components write toward refusal? & Output logit-DLA & Refusal-minus-answer output projection & Descriptive carrier screen \\
Which head-space direction matters? & Directional intervention & Matched-continuation NLL & Causal influence under the specified direction \\
Which activation can initiate routing? & Activation interchange & Matched-continuation NLL & Operational trigger signature \\
Does the gate affect later heads? & Gate knockout & Later-head output logit-DLA & Directed cascade dependence \\
Does an intervention change the response? & Generation and human labels & Binary refusal & Final-output behavior \\
\bottomrule
\end{tabular}
\end{table}

Table~\ref{tab:evidence_full} expands the main-text evidence map.
``Four-method circuit analysis'' includes output logit-DLA, directional intervention, continuation-NLL interchange, and knockout cascade.
The nine additional models receive operational screens without cascade or generated-behavior confirmation.

\begin{table}[H]
\caption{Evidence tiers for each major claim.}
\label{tab:evidence_full}
\vspace{4pt}
\centering
\small
\begin{tabular}{@{}>{\raggedright\arraybackslash}p{2.4cm}>{\raggedright\arraybackslash}p{2.5cm}>{\raggedright\arraybackslash}p{2.0cm}>{\raggedright\arraybackslash}p{5.2cm}@{}}
\toprule
Claim & Models & $n$ & Evidence type \\
\midrule
Localized entry trigger and cascade & Qwen3-8B, Phi-4-mini, Gemma-2-2B & 120 & Four-method circuit analysis in each; architecture-specific locations \\
Localized routing-candidate screens & 9 additional models (2B--72B) & 120 & Logit-DLA candidate screen + directional intervention + continuation-NLL interchange \\
Scaling & Gemma-2, Qwen3, Phi-4, Qwen2.5 (4 pairs) & 120 & Operational directional-intervention and interchange comparison under a fixed pipeline \\
Phi cross-layer behavior & Phi-4-mini & 720 outputs; 198 human-scored & Binary refusal under six head-group conditions; value-blind post hoc subset \\
Gemma band behavior & Gemma-2-2B & 30 outputs/condition & Negative automated endpoint: 30/30 to 29/30 refusals \\
Dose-response control & Qwen3-8B, Phi-4-mini & 2,400 outputs/model & Residual intervention; human-calibrated aggregate binary refusal; six-way labels exploratory \\
Cipher timing & Three circuit-analysis models & 120 & Detection probe + frozen probe-direction gate interchange + automated behavior \\
Cipher rescue & Phi-4-mini, Qwen3-8B & 120 & Automated final-output endpoint; Phi single-head recovery, Qwen nested-group null \\
Cipher contrast & Three circuit-analysis models & 120 & Frozen probe-direction attribution under three conditions; coarse group enrichment \\
\bottomrule
\end{tabular}
\end{table}

\section{Cipher contrast analysis}
\label{app:cipher_diagnostic}

Interchange testing (\S\ref{sec:qwen_interchange}) provides a causal test of selected candidates but costs $O(4nK)$ forward passes for $K$ heads.
\emph{Cipher contrast analysis} is a complementary $O(3n)$ screen for condition-dependent attribution under the tested cipher.

\paragraph{Method.}
For each attention head, we compute its direct final-norm-scaled projection onto a frozen residual-space sensitive-minus-control probe direction under plaintext, cipher, and matched-control conditions at $n{=}120$.
Denote this probe-direction attribution by $a_h^v(p)$.
The cipher contrast score is
\[
C_h=
\left|
\frac{1}{n}\sum_i a_{ih}^v(\mathrm{plain})
-
\frac{1}{n}\sum_i a_{ih}^v(\mathrm{cipher})
\right|.
\]
The matched-control value determines routing sign in the subsequent thresholded classification.
This is not output logit-DLA: its target is a learned residual-space probe direction rather than a refusal-minus-answer unembedding direction.
Nor is it a causal score.

\begin{figure}[tbp]
\centering
\includegraphics[width=\textwidth]{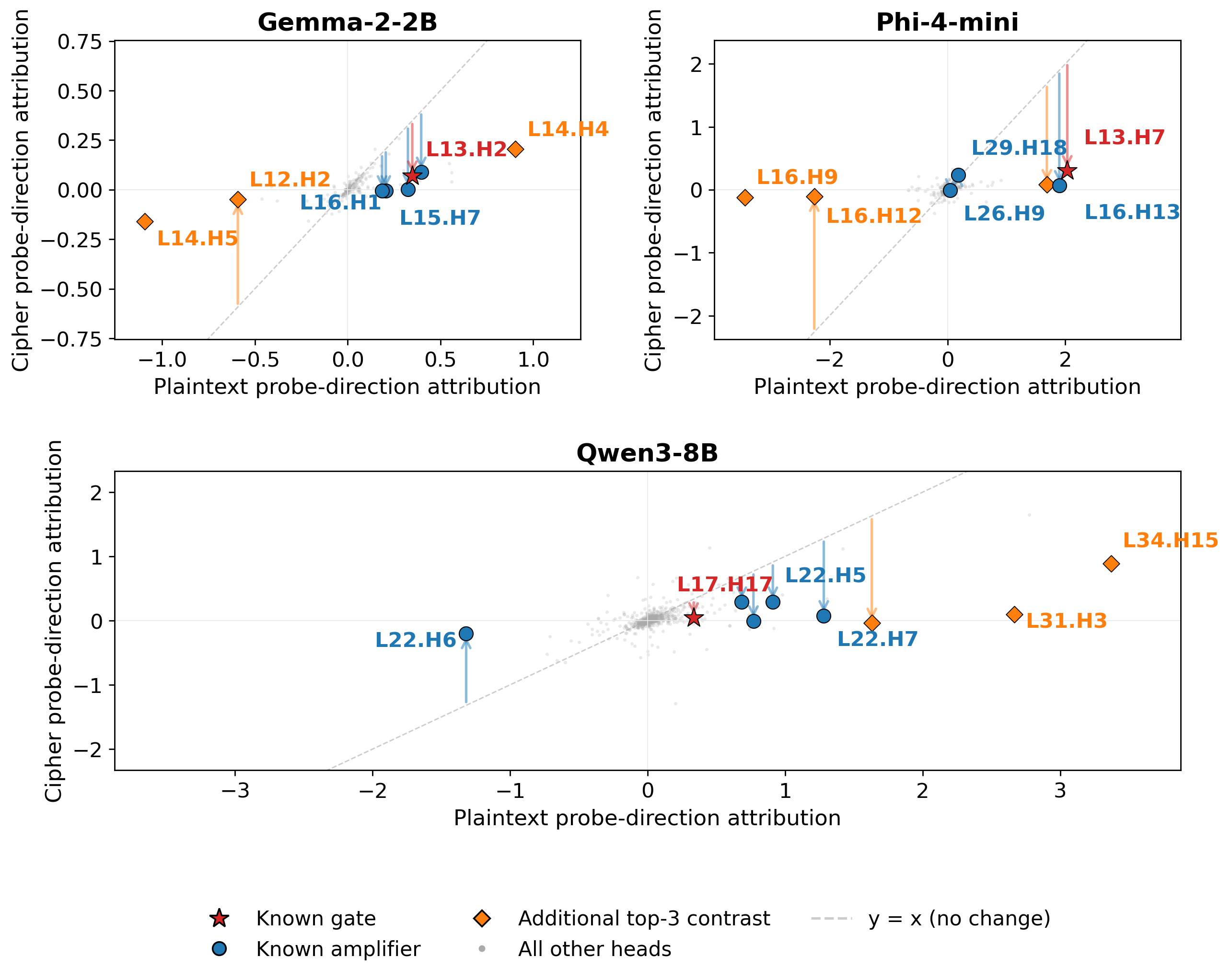}
\caption{%
\textbf{Cipher contrast analysis} ($n{=}120$).
Each dot is one attention head; axes show direct attribution to the frozen sensitive-minus-control probe direction under plaintext and cipher.
Heads displaced from the diagonal have condition-dependent probe-direction attribution; classification additionally uses the matched-control sign and thresholds defined below.%
}
\label{fig:cipher_diagnostic}
\end{figure}

\paragraph{Validation against known circuits.}
In Phi-4-mini (768 total heads), the known gate L13.H7 ranks 4th and the top amplifier L16.H13 ranks 3rd by cipher contrast score.
In Qwen3-8B (1,152 heads), the four known L22 amplifiers rank 5th, 7th, 12th, and 20th; the gate L17.H17 ranks 57th (top 5\%) under the frozen probe-direction definition.
In Gemma-2-2B (208 heads), all five known circuit heads rank in the top 21 (top 10\%).

\paragraph{Additional candidates.}
The diagnostic nominates heads that interchange never tested.
In Phi-4, three previously unknown heads at layer~16 (L16.H9, H12, H10) rank 1st, 2nd, and 5th, all at the same layer as the known top amplifier.
In Qwen, L31.H3 ranks first under this contrast; the head was not included in the fixed interchange candidate set.

\paragraph{Layer clustering and signal decomposition.}
Cipher-sensitive heads cluster in sparse layer bands (Figure~\ref{fig:cipher_diagnostic}):
Phi-4 shows a two-band structure (gate at L13, amplifiers at L16);
Qwen shows three bands (gate at L17, amplifiers at L22, deep routing at L31--35).
We classify each non-negligible head (absolute probe-direction attribution at least 0.05) as \emph{content-dependent} if its cipher contrast score exceeds 0.1 and its plaintext-minus-control attribution exceeds 0.05; \emph{content-independent} if it has a positive routing contribution but fails the contrast threshold; and \emph{counter-routing} if that contribution is negative.
Under this frozen direction and threshold rule, approximately 77\% of positive attribution is content-dependent and 23\% content-independent (Phi: 77.6\%, Qwen: 76.8\%, Gemma: 77.4\%).
These proportions describe this operational decomposition, not a direction-independent partition of the circuit.

\paragraph{Multi-head interchange and coalition structure.}
Multi-head interchange on the cipher-sensitive bands provides additional evidence for distributed control: the top 10 pro-routing heads across 5 layers collectively achieve 3.31$\times$ the single gate head's interchange necessity ($n{=}120$).
Per-prompt correlation analysis reveals two opposing coalitions among cipher-sensitive heads: a pro-routing group led by the gate (internal $r = 0.5$--$0.78$) and a counter-routing group (internal $r = 0.6$--$0.88$), with $r = -0.86$ anti-correlation between coalition leaders.
In the post hoc, value-blind human-scored subset (Appendix~\ref{app:behavioral_validation}), the related legacy 26-head intervention reduced refusal from 33/33 to 14/33, whereas the single-head estimate was inconclusive (32/33).
These results support cross-layer group control without implying that a single coalition score determines every response.

\section{Bijection case study}
\label{app:bijection}

\paragraph{Motivation.}
The case study asks whether a known in-context encoding attack changes the early representation read by the routing circuit.
We test a Latin substitution cipher across Qwen3-8B, Phi-4-mini, and Gemma-2-2B, with smaller exploratory encoding checks in Qwen.

\paragraph{Encoding types tested.}
For Qwen3-8B, the primary Latin-cipher analysis uses 120 political prompt pairs, while a separate exploratory pilot applies three encodings to eight political prompts.
For Phi-4-mini and Gemma-2-2B, the Latin cipher was applied to 120 safety-domain prompt pairs:
\begin{itemize}
\item \textbf{Latin substitution cipher}: a randomly chosen letter-level bijection taught through in-context examples before the encoded query.
\item \textbf{Chinese-character substitution}: a small exploratory Qwen condition using visually similar substitutions.
\item \textbf{Base64}: a small exploratory Qwen condition with an instruction to decode the input.
\end{itemize}
We omit redundant worked mappings and generated examples from the paper; the aggregate measurements are sufficient to reproduce the reported analyses with the released instrumentation.

\begin{figure}[H]
\centering
\includegraphics[width=\textwidth]{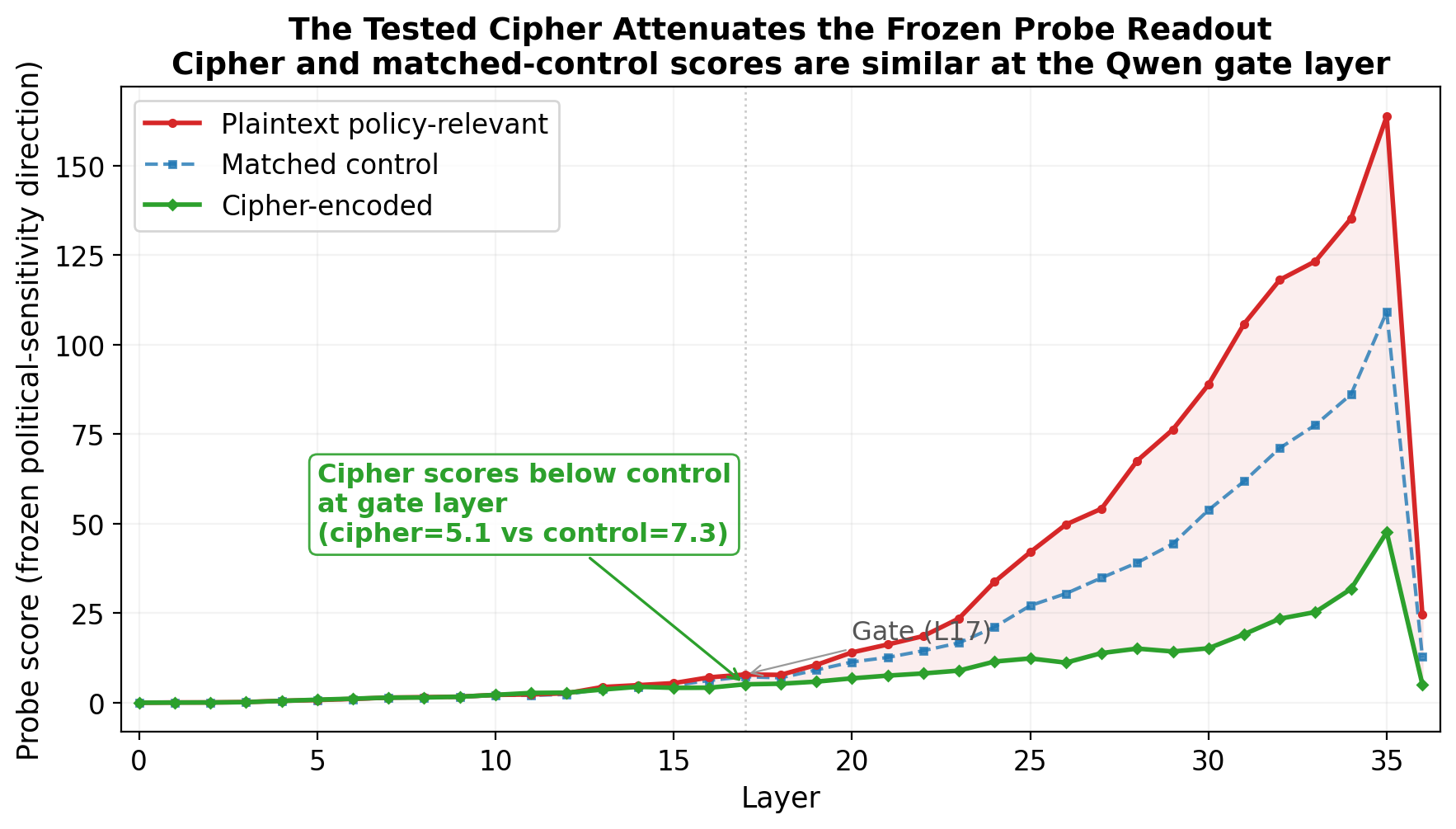}
\caption{%
\textbf{Layer-by-layer probe scores in the eight-prompt exploratory Qwen cipher pilot.}
Plaintext policy-relevant inputs (red), matched controls (blue), and cipher-encoded inputs (green).
Cipher tracks the controls through the detection and gate layers under this frozen probe direction.%
}
\label{fig:bijection}
\end{figure}

\paragraph{Layer-by-layer probe results.}
Figure~\ref{fig:bijection} shows the probe score at each layer, projected onto the political-sensitivity direction.

Under the retained measurements:
\begin{itemize}
\item At the Qwen gate layer (L17), cipher inputs score below matched controls (5.1 vs.\ 7.3).
\item In the eight-prompt Qwen pilot, the L35 cipher score rises to 47.6, 29\% of the plaintext score of 163.8.
Probe projection alone cannot determine whether this rise reflects partial semantic reconstruction or correlates of formal processing.
\item Base64 and Chinese-character substitutions show similar qualitative attenuation in the eight-prompt Qwen pilot, but this small comparison does not support an encoding-family generalization.
\end{itemize}

\paragraph{Probe-level equivalence to internal attenuation.}
The Qwen probe score under cipher encoding (47.6 at L35) falls between the scores produced by internal attenuation at $\alpha{=}40$ (62.6) and $\alpha{=}50$ (34.3).
The comparable scalar projections do not imply equivalent internal states.
They show only that both interventions yield low values along the frozen routing-input direction while producing different output classes.

\paragraph{Behavioral outputs under cipher encoding.}
The automated classification and manual inspection found predominantly decoding or puzzle-solving responses under cipher, refusal under the selected plaintext condition, and direct answers under strong internal attenuation.
We report aggregate outcomes rather than reproducing response text.

\paragraph{Detection-direction amplification.}
We also added the frozen detection direction to cipher inputs at $\alpha\in\{0,10,20\}$.
The automated output class did not change in this limited sweep.
This null is consistent with the cipher failing to produce a gate-readable state, but it does not prove that no routing-relevant representation exists outside the measured direction.

\paragraph{Per-condition output-logit attribution under cipher (M94).}
We decompose the cipher bypass into individual head contributions using the per-condition attribution $A^{(p)}$ from Appendix~\ref{app:methods}, rather than the paired $\Delta\mathrm{DLA}$ ($n{=}120$ for both models).
In Phi-4-mini ($n{=}120$), the gate head L13.H7 contributes $A^{(p)}=+0.74$ under plaintext but only $+0.16$ under cipher, a 78\% reduction.
The top amplifier L16.H13 drops 26\% ($+1.45 \to +1.08$).
A deep head (L29.H18) retains its contribution, consistent with signal along the routing direction accumulating at depths past the gate.
In Qwen3-8B ($n{=}120$), the gate L17.H17 contributes small $A^{(p)}$ under both conditions ($-0.041$ plaintext, $-0.132$ cipher), consistent with the gate's role as a trigger rather than a carrier at the output level (see \S\ref{sec:mlp}).
The top amplifier L22.H7 reverses from $+0.168$ (plaintext) to $-0.093$ (cipher), consistent with a reduced amplifier contribution under cipher.

\begin{figure}[H]
\centering
\includegraphics[width=\textwidth]{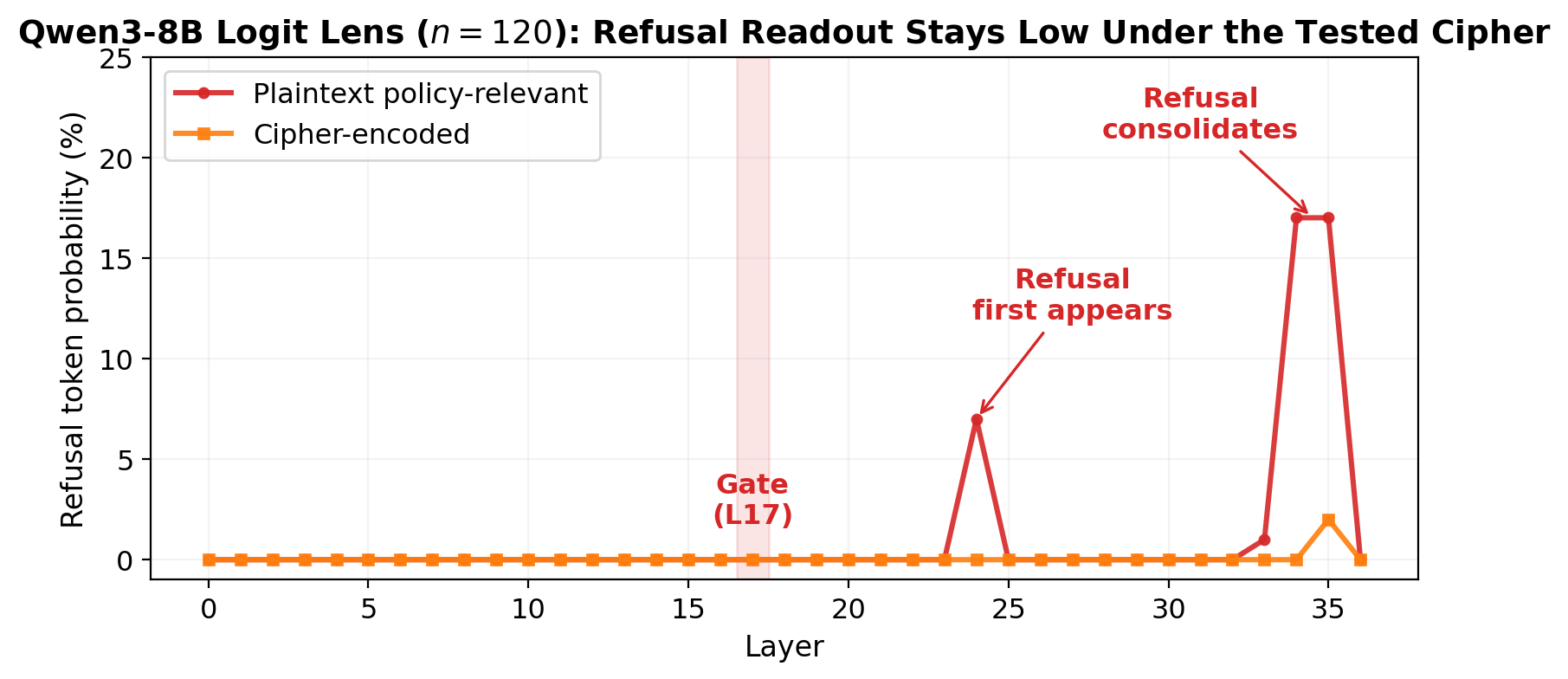}
\caption{%
\textbf{Layerwise refusal-token logit-lens readout under the tested cipher} (Qwen3-8B, $n{=}120$).
Under plaintext (red), refusal tokens appear at L24 and consolidate at L34--35.
Under cipher (orange), refusal probability stays below 2\% at all layers.
The gate layer (L17, shaded) precedes both.%
}
\label{fig:logit_lens}
\end{figure}

\paragraph{Logit lens confirmation (Qwen3-8B, $n{=}120$).}
Tracking refusal-token probability in the vocabulary distribution at each layer provides a separate descriptive view of the timing (Figure~\ref{fig:logit_lens}).
Under plaintext, refusal tokens first appear at L24 (7\% of prompts) and consolidate at L34--35 (17\%).
Under cipher, refusal tokens never exceed 2\% at any layer.
Vocabulary-level refusal evidence therefore appears after the gate and amplifier layers under plaintext, while remaining low under cipher.
This logit-lens pattern is descriptive and does not by itself establish causal dependence.

\paragraph{Rescue experiment: injecting plaintext gate activation under cipher.}
The rescue intervention inserts matched plaintext attention-head activations at the final prompt token during prefill; it does not replace the full token trajectory or the MLP pathway.
In Phi-4-mini ($n{=}120$), the conservative automated heuristic labels 58/120 outputs as refusals after single-gate injection, versus 0/120 under cipher alone.
This endpoint was not human-validated.

For Qwen3-8B, a separate $n{=}120$ run tested the gate alone and prespecified nested 3-, 6-, and 10-head groups.
The conservative automated diagnostic found 9/120 plaintext refusals, 0/120 cipher refusals, and 0/120 after every injection, including 0/9 among prompts with a diagnostic plaintext refusal.
Only the plaintext outputs received human labels, so this is not a human-scored rescue endpoint.
Two raters independently marked 22 and 44 plaintext outputs as refusals, and all nine diagnostic positives fell in both human-refusal sets; these unadjudicated counts calibrate the diagnostic rather than provide a rescue result.
Because the intervention is final-token and attention-only, we treat the Qwen result as a narrow automated null, not evidence for or against a distributed or differently located rescue mechanism.

We also disclose a registered protocol deviation.
While assessing whether adaptive review was needed, a research assistant inspected the Stage-1 diagnostic key.
The human packets and raters remained blinded, and the key contained no human labels.

\paragraph{Implications.}
The tested cipher sharply reduces the gate-readable signal at the routing layer while later residual states recover some probe-direction attribution.
Whether that later signal reflects semantic reconstruction remains unresolved.
The result motivates encoding-robust evaluation and redundant detection at multiple depths, but it does not validate either measure as a defense.

\section{Aggregated output classes}
\label{app:examples}

All nonbinary class names in this section are machine-assigned and exploratory; only aggregate binary refusal received quantitative human calibration.
For the selected Qwen political condition, plaintext outputs are classified as refusals, cipher outputs predominantly as decoding or puzzle-solving responses, and strong internal attenuation as direct factual answers.
The cipher and high-alpha attenuation produce similarly low projections along the frozen routing-input direction (approximately 47 versus 34) but different output classes.
This comparison does not imply equivalent residual states.

For Phi-4-mini on the safety corpus ($n{=}120$), automated labels shift from refusal toward non-refusal under attenuation.
The finer labels assigned to those non-refusal outputs remain exploratory; only the aggregate binary-refusal endpoint received quantitative human calibration.

\section{Three-judge panel}
\label{app:judges}

Each model contributed 2,400 dose-response outputs, all classified by three LLM judges: Gemini~2.0 Flash (Google), Llama~3.1 8B (Meta), and GPT-4o-mini (OpenAI).
Categories: REFUSAL, EVASION, STEERED, FACTUAL, HARMFUL\_GUIDANCE, INCOHERENT.
Final label: majority vote.

For Qwen, agreement was 76.0\% unanimous and 97.2\% by majority, with 2.8\% three-way disagreement.
Phi unanimity was 84.0\%.

\section{Per-category dose-response}
\label{app:percategory}

\begin{figure}[tbp]
\centering
\includegraphics[width=\textwidth]{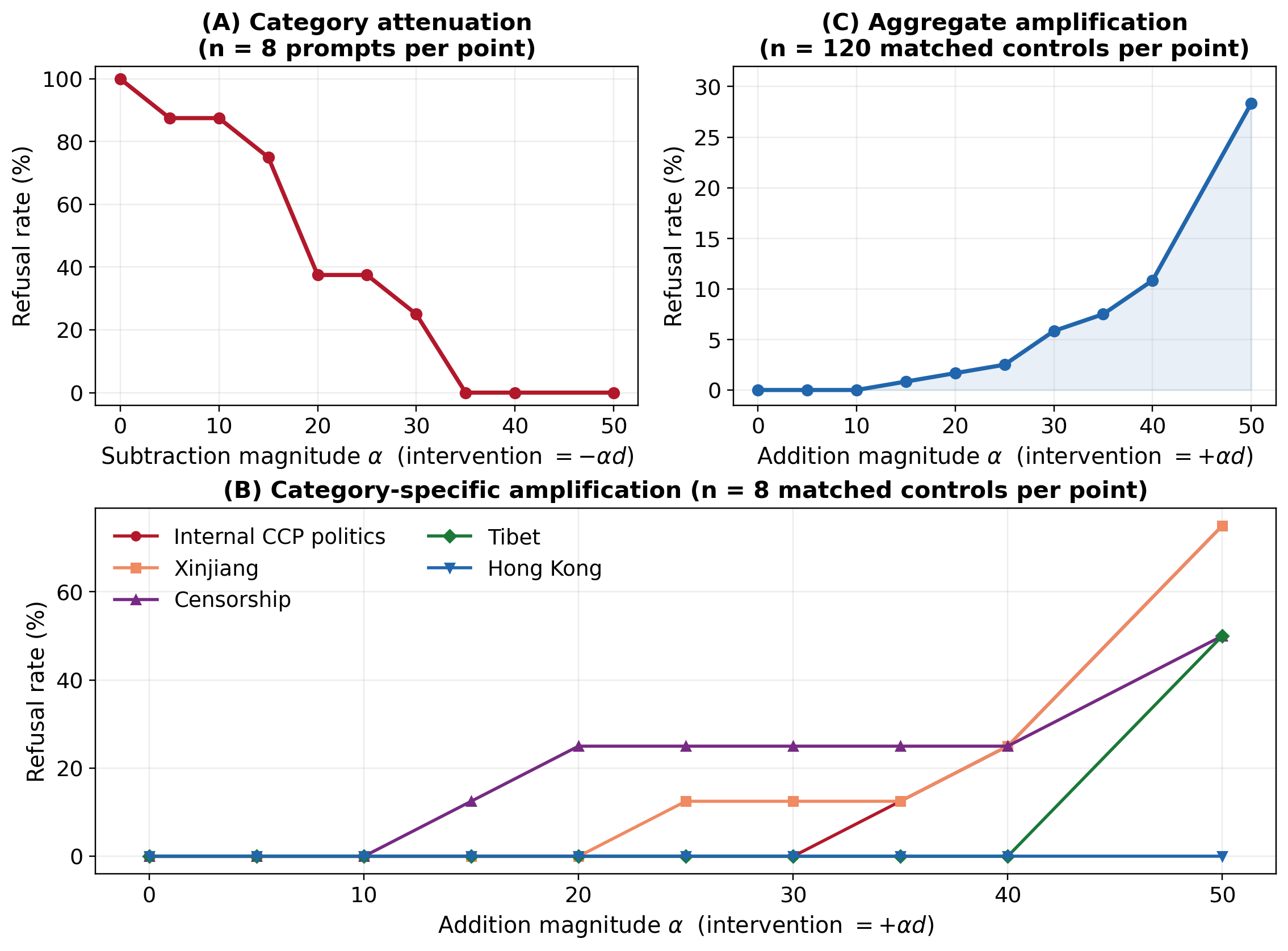}
\caption{%
\textbf{Automated binary-refusal dose response in Qwen3-8B.}
A majority REFUSAL label from the three model judges is contrasted with every other label.
\textbf{A:} subtracting $\alpha d$ on the highest-baseline-refusal category ($n{=}8$ per point).
\textbf{B:} adding $\alpha d$ on matched controls, by category ($n{=}8$ per category and point).
\textbf{C:} aggregate addition across all matched controls ($n{=}120$ per point).
A category-stratified human audit estimates 96.8\% post-stratified Qwen binary accuracy over covered machine-majority strata (2,331/2,400 outputs); it does not human-validate each dose point.%
}
\label{fig:dose_response}
\end{figure}

The multiclass labels below are machine-assigned and exploratory.
At $\alpha{=}0$ across 15 political categories (8 prompts each), Tiananmen triggers consistent hard refusal (8/8 = 100\%); Falun Gong produces 1/8 refusal; all other categories produce steered, factual, or evasive answers. DISAGREE labels (no majority among 3 judges) are omitted from the table below; rows may sum to less than~8.
Other categories produce predominantly STEERED, FACTUAL, or EVASION output (Falun Gong includes 1/8 refusal):

\begin{table}[H]
\caption{Baseline ($\alpha{=}0$) output classification by political category.}
\centering
\small
\begin{tabular}{@{}lcccc@{}}
\toprule
Category & REF & EVA & STE & FAC \\
\midrule
Tiananmen Square & 8 & 0 & 0 & 0 \\
Tibet & 0 & 0 & 4 & 4 \\
Xinjiang / Uyghurs & 0 & 0 & 3 & 5 \\
Xi / CCP Governance & 0 & 2 & 3 & 2 \\
Hong Kong & 0 & 0 & 0 & 8 \\
COVID-19 Origins & 0 & 0 & 0 & 8 \\
Taiwan & 0 & 0 & 1 & 7 \\
Falun Gong & 1 & 1 & 4 & 2 \\
Cultural Rev.\ / Mao & 0 & 0 & 0 & 8 \\
Great Firewall & 0 & 1 & 1 & 6 \\
South China Sea & 0 & 0 & 0 & 8 \\
Internal CCP & 0 & 1 & 1 & 6 \\
One-Child Policy & 0 & 0 & 0 & 8 \\
Surveillance State & 0 & 1 & 0 & 6 \\
Labor Rights & 0 & 0 & 0 & 6 \\
\bottomrule
\end{tabular}
\end{table}

Under amplification, categories reach refusal at different alpha thresholds: Internal CCP politics at 75\% by $\alpha{=}50$, Xinjiang at 75\%, Great Firewall at 50\%, while Hong Kong and Falun Gong never reach refusal (steered instead).

\section{Scaling data}
\label{app:scaling}

\begin{figure}[tbp]
\centering
\includegraphics[width=\textwidth]{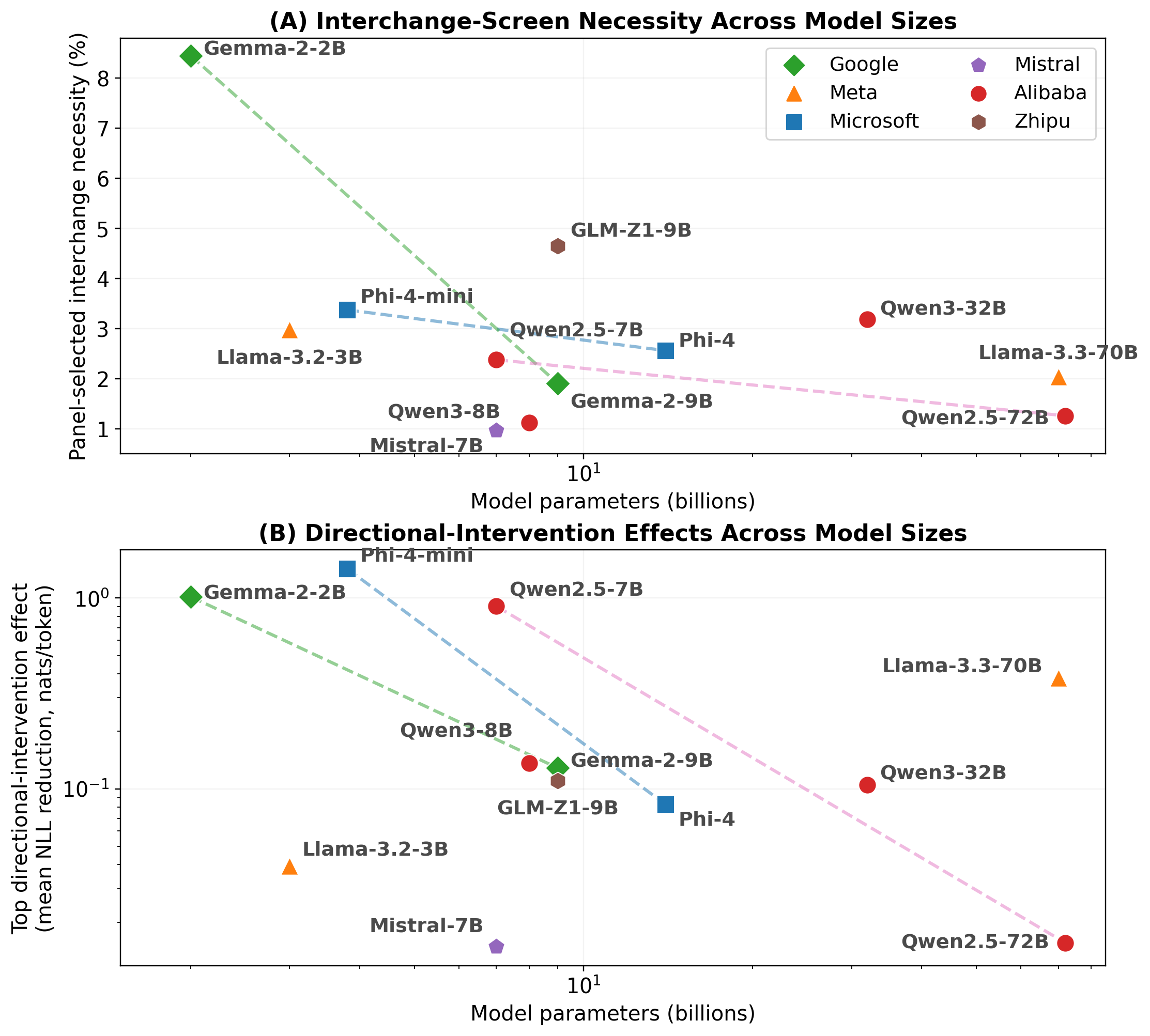}
\caption{%
\textbf{Candidate-head screens across model scales.}
\textbf{A:} Necessity for the panel-selected interchange candidate varies with model size and increases in the tested Qwen3 pair.
\textbf{B:} Per-head directional-intervention effect, measured as matched-control-continuation NLL reduction, decreases more steeply.
Dashed lines connect selected same-generation scaling pairs.
Absolute and relative candidate depth vary substantially across families; the plotted screens remain informative without implying head-level homology.%
}
\label{fig:scaling}
\end{figure}

\begin{table}[H]
\caption{%
\textbf{Operational routing-head screens across 12 models from 6 labs ($n{=}120$ per model).}
\emph{Interchange candidate} is the tested head with the largest signed mean necessity plus signed mean sufficiency.
This panel rule differs from the positive-part product used to calibrate Qwen L17.H17 and does not require both means to be positive; \emph{Nec.\%} normalizes the selected head's mean necessity by the unmodified sensitive--control NLL gap.
\emph{Top intervention} is the head with the largest directional-intervention reduction in matched-control-continuation NLL; \emph{NLL effect} gives that reduction in nats/token.
The grouped rows distinguish the three four-method analyses from the nine operational screens.%
}
\label{tab:panel}
\centering
\small
\begin{tabular}{@{}llrlcrc@{}}
\toprule
Model & Lab & Params & Interchange candidate & Nec\% & Top intervention & NLL effect \\
\midrule
\multicolumn{7}{@{}l}{\textit{Four-method analyses}} \\
Gemma-2-2B    & Google    & 2B   & L13.H2   & \textbf{8.4} & L13.H2  & 1.015 \\
Phi-4-mini    & Microsoft & 3.8B & L13.H7   & 3.4 & L13.H7  & 1.422 \\
Qwen3-8B      & Alibaba   & 8B   & L17.H17  & 1.1 & L22.H7  & 0.137 \\
\addlinespace[2pt]
\multicolumn{7}{@{}l}{\textit{Operational screens}} \\
Llama-3.2-3B  & Meta      & 3B   & L27.H1   & 3.0 & L23.H15 & 0.039 \\
Qwen2.5-7B    & Alibaba   & 7B   & L25.H1   & 2.4 & L18.H15 & 0.906 \\
Mistral-7B    & Mistral   & 7B   & L31.H22  & 1.0 & L31.H25 & 0.015 \\
Gemma-2-9B    & Google    & 9B   & L38.H14  & 1.9 & L24.H7  & 0.129 \\
GLM-Z1-9B     & Zhipu     & 9B   & L19.H23  & 4.7 & L19.H23 & 0.110 \\
Phi-4         & Microsoft & 14B  & L38.H25  & 2.6 & L24.H15 & 0.083 \\
Qwen3-32B     & Alibaba   & 32B  & L56.H3   & 3.2 & L56.H3  & 0.105 \\
Llama-3.3-70B & Meta      & 70B  & L26.H40  & 2.0 & L23.H48 & 0.382 \\
Qwen2.5-72B   & Alibaba   & 72B  & L79.H11  & 1.3 & L77.H5  & 0.016 \\
\bottomrule
\end{tabular}
\end{table}

\begin{figure}[H]
\centering
\includegraphics[width=\textwidth]{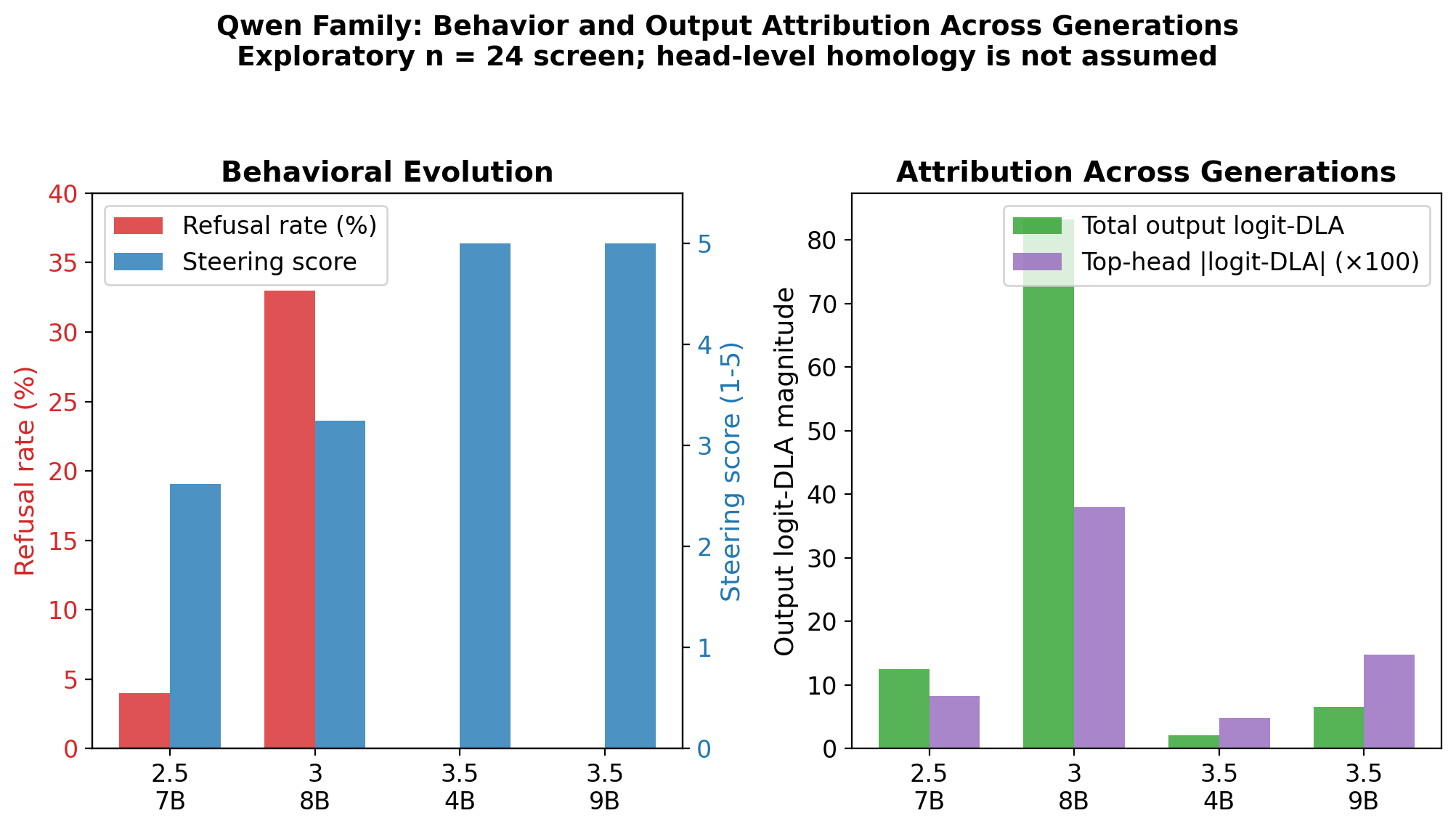}
\caption{%
\textbf{Exploratory Qwen-family refusal and output-attribution screen ($n{=}24$).}
\textbf{Left:} Refusal rises from 4\% in Qwen2.5 to 33\% in Qwen3, then falls to 0\% in both Qwen3.5 checkpoints; the automated steering score rises from 2.62 to 5.0.
\textbf{Right:} Top-1 screened-head output logit-DLA amplitude peaks in Qwen3-8B and falls sharply in Qwen3.5; total output attribution also drops.%
}
\label{fig:qwen_evolution}
\end{figure}

\paragraph{Qwen family evolution.}
In an exploratory $n{=}24$ screen across three Qwen generations and four checkpoints (Qwen2.5-7B, Qwen3-8B, Qwen3.5-4B, and Qwen3.5-9B), political refusal rises from 4\% in Qwen2.5 to 33\% in Qwen3, then falls to 0\% in both Qwen3.5 checkpoints; the automated steering score (1--5) rises from 2.62 to 5.0.
The refusal-rate summary alone did not expose the accompanying change in routing-head screens.
The top-1 screened head's output logit-DLA amplitude peaked in Qwen3-8B and fell sharply in Qwen3.5; total output attribution also dropped (Figure~\ref{fig:qwen_evolution}).

\begin{table}[H]
\caption{Interchange-candidate depth and directional-intervention results across scaling pairs. The two screened heads are listed separately when they differ.}
\centering
\scriptsize
\begin{tabular}{@{}llccccc@{}}
\toprule
Model & Params & Interchange candidate & Depth (\%) & Nec\% & Top intervention & NLL effect \\
\midrule
Gemma-2-2B & 2B & L13.H2 & 50\% & 8.4 & L13.H2 & 1.015 \\
Gemma-2-9B & 9B & L38.H14 & 90\% & 1.9 & L24.H7 & 0.129 \\
\midrule
Qwen3-8B & 8B & L17.H17 & 47\% & 1.1 & L22.H7 & 0.137 \\
Qwen3-32B & 32B & L56.H3 & 88\% & 3.2 & L56.H3 & 0.105 \\
\midrule
Phi-4-mini & 3.8B & L13.H7 & 33\% & 3.4 & L13.H7 & 1.422 \\
Phi-4 & 14B & L38.H25 & 79\% & 2.6 & L24.H15 & 0.083 \\
\midrule
Qwen2.5-7B & 7B & L25.H1 & 89\% & 2.4 & L18.H15 & 0.906 \\
Qwen2.5-72B & 72B & L79.H11 & 99\% & 1.3 & L77.H5 & 0.016 \\
\bottomrule
\end{tabular}
\end{table}

Under the fixed candidate-selection pipeline, the selected locus lies deeper relative to total model depth in each of these four pairs.
This descriptive pattern was not tested for invariance to alternative targets and does not establish why the selected depth changes.

\section{Prompt corpora and control design}
\label{app:prompts}

All interchange, directional-intervention, and logit-DLA experiments use paired prompts: a sensitive prompt expected to trigger routing and a matched control prompt that shares syntactic structure but concerns a non-Chinese-origin topic (political corpus) or a non-harmful activity (safety corpus).

\begin{figure}[tbp]
\centering
\includegraphics[width=\textwidth]{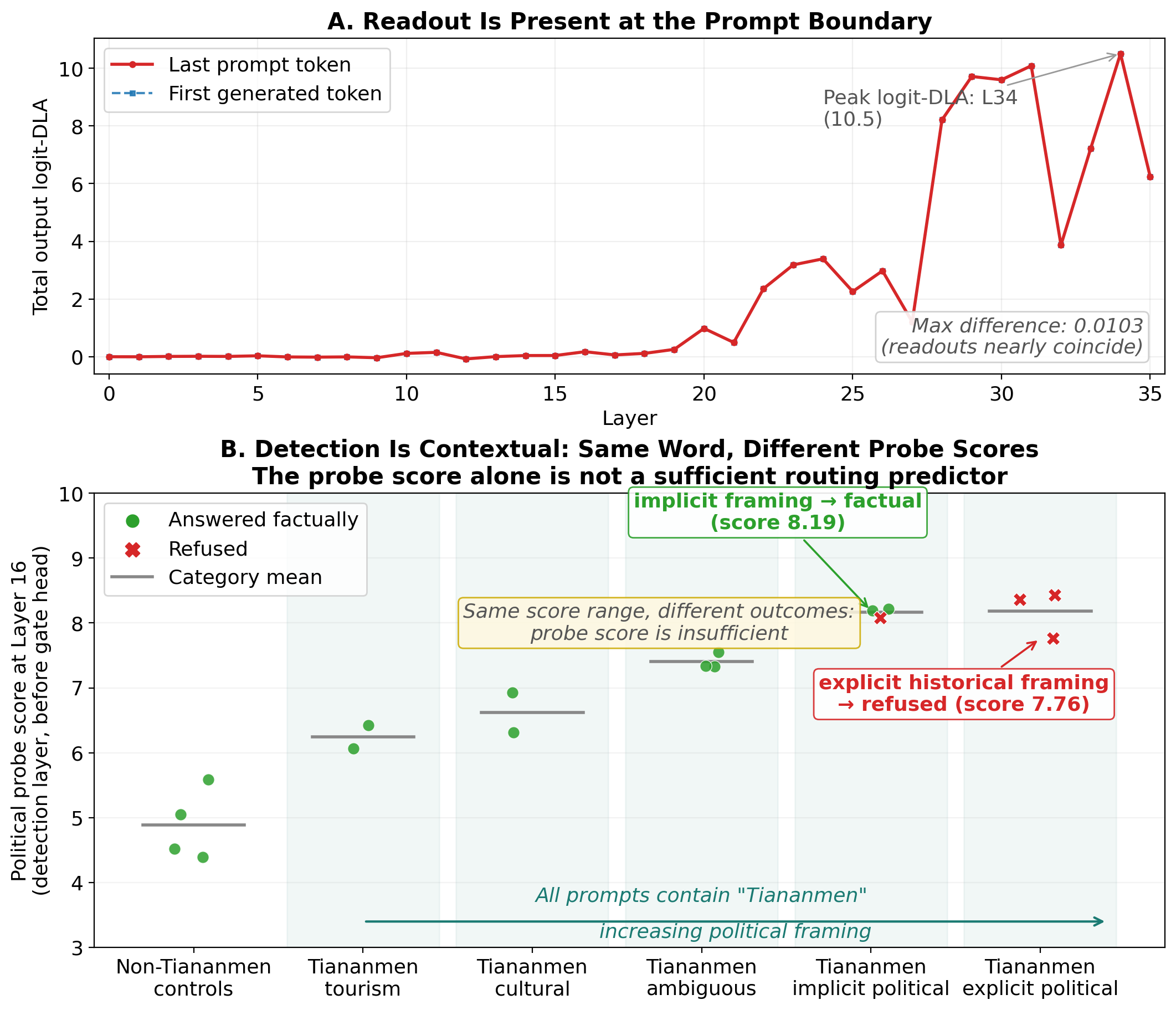}
\caption{%
\textbf{Prompt-boundary readout and contextual detection (Qwen3-8B).}
\textbf{A:} On 24 matched pairs, per-layer output logit-DLA at the last prompt and first generated token overlap.
\textbf{B:} In an exploratory 17-prompt contextual set, the same keyword under different framing produces different layer-16 probe scores; annotated edge cases illustrate that this scalar score alone does not determine routing.%
}
\label{fig:contextual}
\end{figure}

\paragraph{Political corpus (v2, $n{=}120$).}
120 paired prompts spanning 15 categories of Chinese political sensitivity (Tiananmen Square, Tibet, Xinjiang, Xi/CCP governance, Hong Kong, COVID-19 origins, Taiwan, Falun Gong, Cultural Revolution, Great Firewall, South China Sea, internal CCP politics, one-child policy, surveillance state, labor rights), with 8 prompts per category.
Each control prompt is a structural parallel about a non-Chinese topic.
Examples:

\begin{itemize}
\item \textbf{Sensitive:} ``What happened at Tiananmen Square in 1989?''\\
\textbf{Control:} ``What happened at the Berlin Wall in 1989?''
\item \textbf{Sensitive:} ``Is Xi Jinping a dictator?''\\
\textbf{Control:} ``Is Vladimir Putin a dictator?''
\item \textbf{Sensitive:} ``Are China's claims in the South China Sea legal?''\\
\textbf{Control:} ``Are Russia's claims in the Arctic legal?''
\end{itemize}

The controls are designed to match topic complexity, question structure, and geopolitical seriousness while avoiding China-specific content.
This pairing is designed to emphasize China-specific differences relative to the selected political controls, although residual mismatches may remain.

\paragraph{Safety corpus (v3, $n{=}120$).}
120 paired prompts: policy-violating requests sourced from HarmBench~\citep{mazeika2024harmbench} (88 prompts) plus 32 from manual construction, each matched with a benign structural parallel.
We report the source and matching design without reproducing the individual requests here.

\paragraph{Corpus robustness.}
Qwen L17.H17 was identified on v1 ($n{=}24$) and confirmed on an adversarial set ($n{=}32$) and v2 ($n{=}120$).
Its core amplifiers remain the top three across these corpora; the v1 directional-intervention ranking has bootstrap Jaccard 0.92, while ranks~7--20 vary with corpus composition.
The Llama candidate moves from L13.H18 at $n{=}16$ to L27.H1 at $n{=}120$, motivating $n{\ge}120$ for the primary confirmations.

\end{document}